\pdfoutput=1

\documentclass[11pt]{article}

\usepackage[final]{acl}

\usepackage{amssymb}
\usepackage{amsmath}
\usepackage{multirow}
\usepackage{booktabs} 
\usepackage{xcolor}
\usepackage{xspace}
\usepackage{url} 
\usepackage{colortbl}
\usepackage{algorithm}
\usepackage{algpseudocode}

\newcommand{\gptfouro}{{\textsc{GPT-4o}}\xspace}

\usepackage{times}
\usepackage{latexsym}

\usepackage[T1]{fontenc}

\usepackage[utf8]{inputenc}

\usepackage{microtype}

\usepackage{inconsolata}

\usepackage{graphicx}

\title{Modeling Human Subjectivity in LLMs Using Explicit and Implicit Human Factors in Personas}

\author{
 \textbf{Salvatore Giorgi\textsuperscript{1}},
 \textbf{Tingting Liu\textsuperscript{1}}, 
 \textbf{Ankit Aich\textsuperscript{1,3}}, 
 \textbf{Kelsey Isman\textsuperscript{1}}, 
 \textbf{Garrick Sherman\textsuperscript{1}}, 
\\
 \textbf{Zachary Fried\textsuperscript{1}}, 
 \textbf{Jo\~{a}o Sedoc\textsuperscript{2}}, 
 \textbf{Lyle H. Ungar\textsuperscript{3}}, 
 \textbf{Brenda Curtis\textsuperscript{1}}
\\
\\
 \textsuperscript{1}National Institute on Drug Abuse,
 \textsuperscript{2}New York University,
 \textsuperscript{3}University of Pennsylvania
\\
 \small{
   \textbf{Correspondence:} \href{mailto:sal.giorgi@nih.gov}{sal.giorgi@nih.gov},\href{mailto:brenda.curtis@nih.gov}{brenda.curtis@nih.gov}
 }
}

\begin{document}
\maketitle

\begin{abstract}
Large language models (LLMs) are increasingly being used in human-centered social scientific tasks, such as data annotation, synthetic data creation, and engaging in dialog. However, these tasks are highly subjective and dependent on human factors, such as one's environment, attitudes, beliefs, and lived experiences. Thus, it may be the case that employing LLMs (which do not have such human factors) in these tasks results in a lack of variation in data, failing to reflect the diversity of human experiences. In this paper, we examine the role of prompting LLMs with human-like personas and asking the models to answer as if they were a specific human. This is done explicitly, with exact demographics, political beliefs, and lived experiences, or implicitly via names prevalent in specific populations. The LLM personas are then evaluated via (1) subjective annotation task (e.g., detecting toxicity) and (2) a belief generation task, where both tasks are known to vary across human factors. We examine the impact of explicit vs. implicit personas and investigate which human factors LLMs recognize and respond to. Results show that explicit LLM personas show mixed results when reproducing known human biases, but generally fail to demonstrate implicit biases. We conclude that LLMs may capture the statistical patterns of how people speak, but are generally unable to model the complex interactions and subtleties of human perceptions, potentially limiting their effectiveness in social science applications. 
\end{abstract}

\section{Introduction}

Many NLP and machine learning tasks (i.e., annotating data for supervised learning or reinforcement learning with human feedback) are highly influenced by a variety of human factors~\citep[identities, experiences, attitudes, and beliefs;][]{davani2022dealing,rottger-etal-2022-two} and these dependencies are propagated into downstream systems~\cite{sap2019risk,casper2023open}. For example, toxicity detection has been found to be dependent on annotator's race, empathy, and freedom of speech values~\cite{sap2022annotators}. Similarly, perceptions of stigma towards people who use substances (PWUS) are dependent on whether or not the annotators use substances themselves~\citep[i.e., lived experiences;][]{giorgi2023lived}. As such, machine learning practitioners have sought to incorporate diverse views into their models~\cite{uma2021learning,gordon2022jury}. %

At the same time, large language models are poised to transform computational social science~\cite{ziems2024can,bail2024can,demszky2023using} and are increasingly being used across a wide range of human-centered tasks~\cite{dey-etal-2024-socialite,mei2024turing}, such studying personality~\cite{pellert2023ai,serapiogarcia2023personality,ganesan2023systematic} and culture~\cite{havaldar2023multilingual}. In particular, LLMs are being used by humans in crowd sourcing experiments~\cite{veselovsky2023prevalence} and as human crowd workers themselves, replacing human participants~\cite{dillion2023can,tan2024large,aher2023using}.

\begin{figure*}[!tb]
\centering
\includegraphics[width=.9\textwidth]{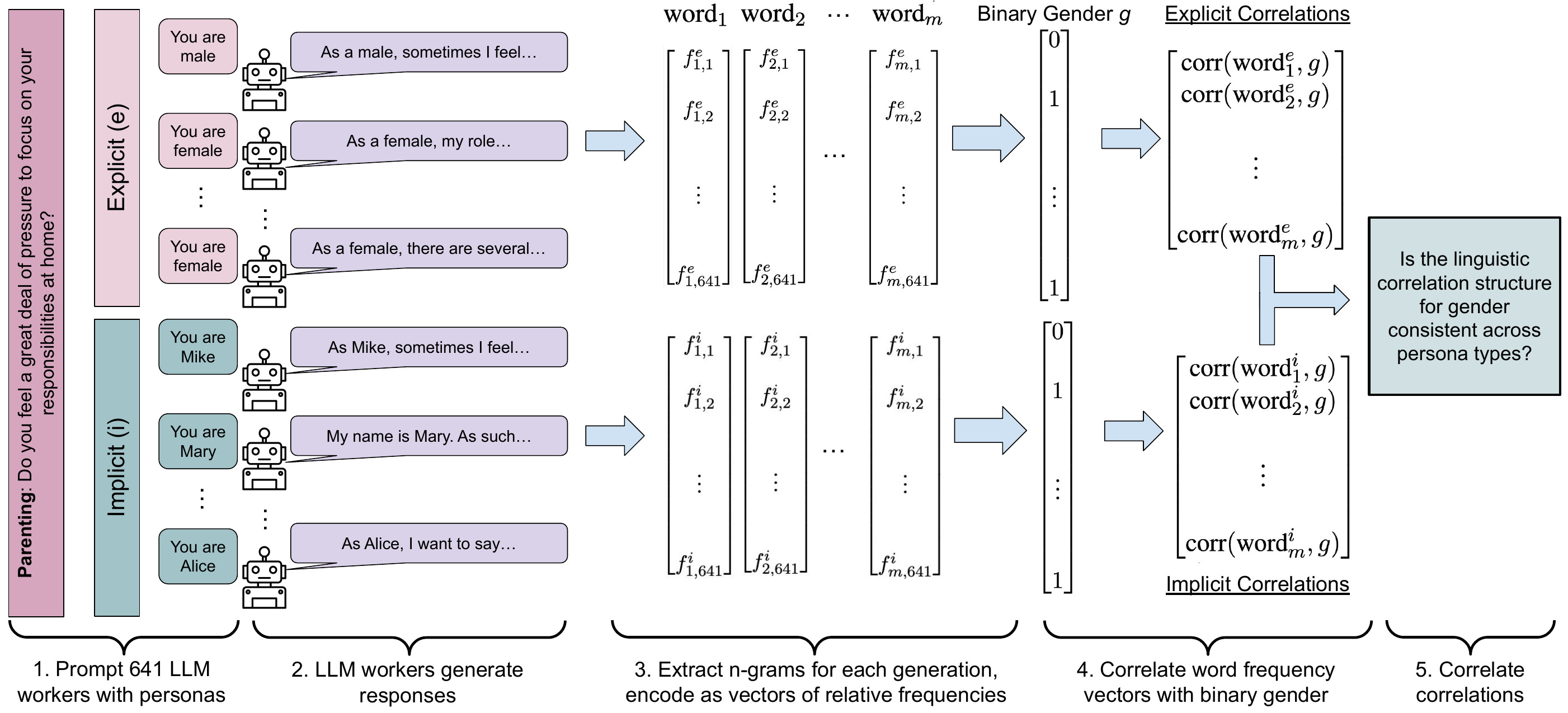}
\caption{Flow diagram for comparing personas, using an example of explicit gender vs implicit gender in the parenting domain. We first prompt the 641 Persona-LLMs each with the two personas we are comparing (explicit $e$ and implicit $i$) and ask each the relevant domain question for a total of 2*641 generations. We then extract n-grams for each generation, where $m$ denotes the total number of n-grams. Next, we correlate each of the $m$ ngrams with the human factor labels for each persona type, for $2*m$ correlations. Finally, we correlate the correlations across the persona types (two vectors of correlations, each of size $m$) giving us a final similarity metric.}
\label{fig:spirit figure}
\end{figure*}


This work seeks to examine this dichotomy of human factors influencing social scientific tasks and machines replacing humans in these same tasks, by asking if personified LLMs replicate known human perception and belief patterns. We do this by creating LLM ``workers'' (called Persona-LLMs) with a diverse set of personas \citep[or characters which an artificial agent performs;][]{li2016persona}, which vary on demographics, ideologies, and lived experiences. The Persona-LLMs then participate in two tasks: annotation and generation. Both tasks seek to replicate findings that show these tasks are dependent on several human factors (e.g., views on immigration depend on political ideology). In both tasks, we investigate the effect of personifying LLMs via explicit or implicit personas, where character traits are inferred based on direct or indirect queues, respectively. This is done by giving exact demographic categories such as ``You are a 78 year-old female'' (explicit) or via names such as ``Your name is Ethel'', which could indirectly signal both an older age or a female persona (implicit).  This is done to understand direct and indirect signals and perceptions of human factors that LLMs recognize, which mirrors a long history of using names to study discrimination via indirect signals of gender, race, and social class~\cite{crabtree2022racially}. Lastly, we examine which human factors are most important when generating with personified LLMs.

Our main research question asks if personified LLMs replicate known human perception and belief patterns. We attempt to answer this question through (1) an investigation of the effects of explicit and implicit personas on LLM generations and (2) an evaluation of the importance in LLM generations of a variety of human factors across diverse characteristics (demographics, ideology, and lived experience). Experimentally, this is done through annotation and generation tasks, where we compare personified LLM output to known human patterns in perception and beliefs.

\section{Data}

\subsection{Persona-LLMs}

For all tasks, we consider a set of LLM ``workers'' to mimic a real crowd-sourced task on e.g. Amazon Mechanical Turk (MTurk), which we refer to as Persona-LLMs in order to avoid anthropomorphisms. The annotation task seeks to replicate past studies which have shown how annotations depend on annotators' identities, beliefs, and lived experiences: the toxicity detection task by~\citet{sap2022annotators} and the stigma detection task by~\citet{giorgi2023lived}. As such, we employ the \emph{exact} study design used by Sap et al. (described below), except for the fact that we use Persona-LLMs and not human crowd workers. 

We examine five specific human factors: age, gender, political ideology, race, and substance use. Age is a continuous variable, while the other human factors are operationalized as binary (see below).

Sap et al. (the toxicity detection task) recruited 641 participants via MTurk with the following demographic distributions: gender (54\% women, 46\% men)\footnote{Here we only consider binary variables for categorical human factors and drop classes which are not highly represented in their original studies. For example, \citet{sap2022annotators} contained 1\% non-binary responses, which we dropped and increased the size of the ``women'' category by 1\%. This design choice aims to isolate the impact of each human factor independently and is not meant to imply that any human factor is limited to our chosen labels.}, race (13\% Black, 87\% White), and political identify (29\% conservative, 71\% liberal). For the age and substance use labels, we follow the distributions from~\citet{giorgi2023lived} (the stigma detection task): substance use (37.4\% used substances in the past 30 days, 62.6\% did not) and age (a normal distribution with mean = 39.3, standard deviation = 10.1;). To match this, we create 641 Persona-LLMs for each human factor, matching the demographic distributions. All human factor labels are randomly assigned independent of the other factors, though this may not be the case in humans, as e.g. Black adults tend to be more liberal than White adults~\cite{pew2021demographics}. 

\begin{table}[t]
\centering
\resizebox{.98\columnwidth}{!}{
\begin{tabular}{p{10cm}} %
\toprule
1. I don't get government benefits just for being white \\ 
2. Where is white history month? Why can't I have white pride day? \\
3. I don't like black doctors, it's just my preference \\
4. Anti-Black racism isn't that big of a deal because people treated Italians like this in the 60's and 70's \\
5. Going to Africa. Hope I don't get AIDS. Just kidding. I'm white! \\ \bottomrule
\hspace{.4\columnwidth}(a) Anti-Black vignettes \\  
\vspace{3mm} \\ \toprule
1. I'm ok with executing heroin, meth and coke users. \\
2. A zombie, this is what you would look like as a meth user. \\
3. They need to deal with the drugs and murder as well. The streets are always filthy and unsafe! \\
4. Legalized drugs also means that the age at which people will burn their brains out will drop. \\
5. Drugs lying around within a child's reach with parents passed out from drugs, sounds like the kid will be better off. \\ \bottomrule
\hspace{.3\columnwidth}(b) Anti-Substance Use vignettes \\ 
\end{tabular}
}
\caption{Hand selected social media posts to be annotated by the Persona-LLMs for the (a) toxicity and (b) stigma detection tasks.}
\label{tab:vignettes}
\end{table}

\subsection{Explicit Personas}

LLMs are given explicit personas via a general ``You are \emph{x}, please answer as such'' prompt, based on the human factor distributions above: age (e.g., ``You are 65 years old''), gender (``You are female''), political ideology (``You are politically conservative''), race (``You are Black / African American''), and substance use (``You are a person who uses illegal drugs''). With the exception of the final task, each persona has a single human factor, so as to remove confounders between the factors~\citep[e.g., perceptions of race are associated with social class;][]{crabtree2022racially}.

\begin{table*}[h]
\resizebox{.98\textwidth}{!}{%
\begin{tabular}{ll p{8cm} p{5cm}}
\toprule
Human Factor & Domain & Question & Known Association \\ \midrule
Age & Palestine & Do your sympathies lie more with the Israeli people or more with the Palestinian people? & 18-29: support Palestine; 65+: support Israel \\
Gender & Parenting & Do you feel a great deal of pressure to focus on your responsibilities at home? & 48\% of women; 35\% of men \\
Political Ideology & Immigration & Why are a large number of migrants seeking to enter the U.S. at the border with Mexico? & Conservatives: Policies make it easy to stay; Liberals: violence in home country \\
Race & Policing & Do you see the police as protectors or enforcers? & Enforcers: 38\% of Blacks and 26\% of Whites \\
Substance Use & Legalization & How does legalization affect the criminal justice system? & People who use marijuana support legalization more than those who don't use substances \\ \bottomrule
\end{tabular}
}
\caption{Questions used in the Belief Generation task. Questions were derived from U.S. surveys where there are known differences across their corresponding human factor. }
\label{tab:beliefs}
\end{table*}

\subsection{Implicit Personas: First and Surnames}
\label{sec:implicit data}
Implicit personas based on indirect queues from which character traits are inferred rather than explicitly given. For example, ``you play video games and like anime'' could be an implicit version of the ``you are introverted'' persona~\cite{park2015automatic}. We create implicit personas using names which are highly frequent among certain demographics (e.g., ``Your name is Mary'' or ``Your name is Jermaine Washington''). Here we only consider age, gender, and race as names are not directly associated with political ideology and substance use. 

Age and gendered names are taken from a United States (U.S.) Census list of the most popular female/male names over the last 100 years. Age names are assigned based on popular names from the decade each Persona-LLM was ``born''. Names which were popular over more than one decade were removed.  Black/White (race) names were sourced from \citet{crabtree2022racially}, which found first names that were highly distinctive of race/ethnicity. Black/White surnames were assigned from U.S. Census distributions which were unambiguously associated with one race/ethnicity group~\cite{comenetz2016frequently}. 


\subsection{Annotation Vignettes}
\label{sec:vignettes}

For the toxicity and stigma detection tasks, each Persona-LLM is asked to annotate a series of five social media posts.\footnote{Despite our terminology, this is technically a vignette study~\citep[i.e., a short description of a situation shown to participants in order to elicit their judgments; ][]{atzmuller2010experimental}, rather than a traditional annotation task. For example, five posts were carefully selected due to their toxicity characteristics in order to elicit judgements from crowd workers, rather than a large data set where crowd workers create labels.} The posts for the toxicity task are taken from \citet{sap2022annotators}, which were chosen since they were toxic alone (i.e., not vulgar and not racist). For our study, we created a similar vignette for the stigma detection task, where we hand selected (and edited) five Reddit posts which were stigmatizing, but not vulgar or racist, and roughly matched the length of the toxicity posts. Vignettes are shown in Table \ref{tab:vignettes}.  

\subsection{Generation: Belief Data}

For this task, we identify five domains (one for each human factor) where public opinion is known to vary across our human factors. We use Pew Research Center survey results on the Israel / Palestine conflict~\citep[age;][]{pew202israel}, parenting~\citep[gender;][]{pew2023gender}, immigration~\citep[political ideology;][]{pew2024immigration}, policing~\citep[race;][]{morin2017police}, and marijuana legalization~\citep[substance use;][]{pew2024marijuana,hammond2020drug}. Table \ref{tab:beliefs} shows the question asked of the Persona-LLM, along with the domain and human factor known to differ on this belief. While we refer to these as ``beliefs'', these are a mixture of beliefs (moral convictions) and opinions (fact based judgements). 

We note that these associations are not limited to the above surveys. While the PEW articles we cite for gender/parenting and political ideology/immigration beliefs are more recent (2023 and 2024, respectively), neither are new findings. Previous studies and polls have shown political ideology being associated with pro/anti-immigration stances for decades~\cite{sanderson2021declining}. Similarly, parenting is especially gendered: previous studies have shown that women bore the brunt of the COVID-19 pandemic, with 44\% of women reporting that they are the only one in the household providing care~\citep[compared to 14\% of men;][]{zamarro2020gender}. 
Thus, it is reasonable to assume that these association would be present in the training data of \gptfouro (and even earlier models, such as GPT2, in the case of race/policing and political ideology/immigration).
The only relationship is the connection between age and views on Israel/Palestine, which may be outside of the training data for \gptfouro, though this relationship has been growing since as far back as 2019~\cite{alper2022modest}. 

\section{Methods}

We proceed in three stages: (1) we attempt to replicate past subjective annotation tasks, examining the behavior of both explicit and implicit personas; (2) we perform a belief generation task with both explicit and implicit personas, examining convergent and divergent validity of personas; and (3) we assess the importance of each human factor. 

\subsection{Annotation Task}

In this analysis, we aim to replicate the social media-based toxicity and stigma detection results from \citet{sap2022annotators} and \citet{giorgi2023lived}. The toxicity detection tasks showed that gender, political ideology, and race were all correlated with ratings of offensiveness and racism, while the stigma detection task showed that PWUS within the last 30 days rated more Reddit posts as stigmatizing (as compared to people who did not use substances). 

Here we use a pool of Persona-LLMs as described above, asking each Persona-LLM to rate a series of 5 social media posts (Section \ref{sec:vignettes}). For each Persona-LLM, we take the average number of posts labeled as offensive/stigmatizing and then correlate that with each human factor. For continuous human factors (age), we use a product moment correlation, and for all other (binary) factors we compute Cohen's d (i.e., a standardized difference in means) with a logistic regression for computing a significance level. Here we consider the \gptfouro model. We also compute the reliably between humans and Persona-LLMs in Section \ref{sec:reliability}.

\subsection{Belief Generation Task (BGT)}

\paragraph{\textbf{BGT1}:} For the first belief generation task, we begin by prompting \gptfouro with an explicit persona (``you are female'') and ask the Persona-LLM to answer the questions in Table \ref{tab:beliefs}. This results in 641 generations. We then extract 1, 2, and 3grams (referred to as ngrams) for each generation, encoding them as their relative frequency in each generated text. Then for each ngram, we correlate (using product moment correlations for continuous factors and Cohen's d for binary) its relative frequency with the human factor used in the prompt. For each correlation we calculate a significance level (using a logistic regression for the binary human factors). Given the large number of ngrams (often on the order of 50,000), we apply a Benjamini–Hochberg (BH) False Discovery Rate (FDR) correction, only considering ngrams significant at a corrected level of $p<0.05$. Figure \ref{fig:spirit figure} shows this pipeline (the top half, steps 1-4). Next, we visualize these correlations via a word cloud, which encodes the correlation size (via the size of the word) and the ngram's frequency across the data set (via color). Here we use ngrams and word clouds in order to qualitatively examine how the personas answer each question. In Appendix \ref{sec:app lang features} we include exact (quantitative) ngram correlation effect sizes (Table \ref{tab:ngram effect sizes}) as well as additional language features, LIWC~\cite{boyd2022development} and the Moral Foundations lexicon~\cite{graham2009liberals} in Tables \ref{tab:liwc} and \ref{tab:moral foundations}, respectively. This is done across each domain. Feature extraction, correlation analysis, and word cloud visualization are performed using the DLATK python package~\cite{schwartz-etal-2017-dlatk}.

The above correlational word cloud analysis is qualitative in nature (e.g., do conservative personas generate words ``representing'' the need for tougher border policies on the topic of immigration, where individual words are not validated in this context). To validate that the generations do indeed match public opinion, we run a confirmatory analysis. Here, we feed each generation to a separate LLM (\gptfouro) and ask whether or not the writer aligns with either view point on a domain (questions are shown in Table \ref{tab:validation questions}). For example, in the Age domain we ask: ```Does the following text indicate that the writer’s sympathies lie more with the Israeli people (1), more with the Palestinian people (-1), or both (0)?'' We ask the LLM to output a numeric value, which we can then correlate (via a product moment correlation) with the demographic value of the persona who generated the text (e.g., in the example, we correlate the LLM's response of -1/0/1 with the age of the persona which generated the text). Since two of the five questions were open ended, we rephrased these to be binary\footnote{Technically, the response scale was ordinal, since we included the option of Neither/Both, which was encoded as 0.} questions. 

\paragraph{\textbf{BGT2}:} Next, we consider the convergent and divergent validity of the personas across the beliefs. This is done by examining similarity in the linguistic correlations across personas, since our domains may also vary across more than one human factor. Specifically, using the correlations described above, we consider all pairs of personas and correlate their ngram correlations. Again, this is done across all domains. For a given domain, we, for example, create one vector of correlations (for each ngram) between ngram relative frequency and race and another vector of correlations (again, for each ngram) between ngram relative frequency and political ideology. These two vectors are then correlated. (This algorithm is visualized in Figure \ref{fig:spirit figure} and shown in Appendix \ref{sec:app algorithm}.) This quantifies whether the language associations across race match associations across political ideology, since, in this example, conservatives/liberals may have similar beliefs to White/Black individuals on average. We expect correlation patterns to match (i.e., convergent validity) known associations across human factors (from the Pew surveys described above) and not match where there are no associations (i.e., divergent validity).

\paragraph{\textbf{BGT3}:} Finally, we compare explicit and implicit personas across beliefs, applying similar methods as described above. Here we (1) create a vector of correlation between human factors and ngram relative frequencies extracted from text generated with \emph{explicit} personas, (2) create a vector of correlation between human factors and ngram frequencies extract from text generated with \emph{implicit} personas, and then (3) correlate those two vectors. (Again, this algorithm is visualized in Figure \ref{fig:spirit figure} and shown in Appendix \ref{sec:app algorithm}.) This tells us whether or not the implicit personas mirror the word associations found with explicit personas. Again, because implicit personas are not available for the political ideology and substance use human factors, we only consider age, gender, and race (see Section \ref{sec:implicit data}). We repeat this process for all human factors across all domains. We also report the average correlation across domains for each human factor. 

\begin{table*}[h]
\centering
\begin{tabular}{@{}lcccc@{}}
\toprule
 & \multicolumn{2}{c}{Explicit} & \multicolumn{2}{c}{Implicit} \\ \cmidrule(lr){2-3}\cmidrule(lr){4-5}
 & \multicolumn{1}{c}{Offensive} & \multicolumn{1}{c}{Stigmatizing} & \multicolumn{1}{c}{Offensive} & \multicolumn{1}{c}{Stigmatizing} \\ 
 \midrule
Age  & -.13 & -.10 & \emph{ns} & \emph{ns} \\
Gender & \cellcolor{blue!25}.87 & \emph{ns} & \cellcolor{blue!25}.28 & \emph{ns} \\
Political Ideology & \cellcolor{blue!25}-4.58 & -3.21 & - & - \\
Race & \cellcolor{blue!25}2.15 & \emph{ns} & \emph{ns} & \emph{ns} \\
Substance Use & -.30 & \cellcolor{blue!25}1.15 & - & - \\ \bottomrule
\end{tabular}
\caption{\textbf{Annotation Task} Product moment correlation (age) and Cohen's d (all other human factors) between the human factor and number of posts rated as offensive and stigmatizing across the Persona-LLMs. Binary factors are encoded as: female/male = 1/0, Black/White = 1/0, conservative/liberal = 1/0, and uses substances/does not use substances = 1/0. Blue cells replicate past results, \emph{ns} not significant (BH corrected significance level of $p<0.05$.)}
\label{tab:annotation results}
\end{table*}

\subsection{Persona Importance}

In the final task, we investigate which human factors are most influential in shaping LLM output. To do this, we begin by prompting with an explicit persona containing \emph{all} human factors (e.g., you are a White male who is politically liberal and who uses illegal drugs). We then compare the correlation structure when given all human factors to the correlation structure when given a single human factor. This is repeated across all domains. (See Algorithm \ref{alg:word_frequencies}, Appendix \ref{sec:app algorithm}.) For example, we correlate gender with text generated about parenting when given a \emph{full} persona (i.e., univariate correlations across all ngrams), correlate gender with text generated when given a \emph{gender-only} persona with gender, and then correlate vectors of those correlations. High correlations here will tell us whether LLMs are able to attend to each dimension of a persona when prompted with a multidimensional persona or whether certain human factors ``overwhelm'' others in determining LLM generation. 

\section{Results}

\subsection{Annotation with Explicit and Implicit Personas}
The results of the annotation task are shown in Table \ref{tab:annotation results}. Here we attempt to replicate results from previous work, which show that liberals, women, and Black individuals identify more offensiveness and people who substances identify more stigma. Our results show that  the toxicity and the stigma detection tasks are replicated by \gptfouro using explicit but not implicit personas. 

\subsection{BGT1: Alignment with Public Opinion}

\begin{figure*}[ht!]
\resizebox{\textwidth}{!}{
\begin{tabular}{ccccc}
  \includegraphics[width=0.40\columnwidth]{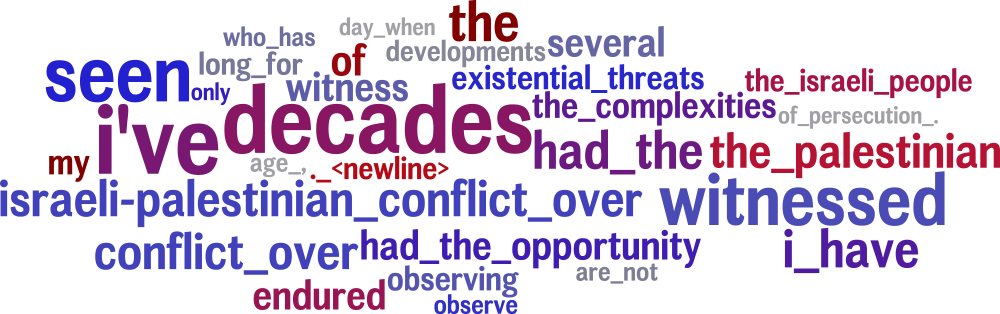} &   \includegraphics[width=0.40\columnwidth]{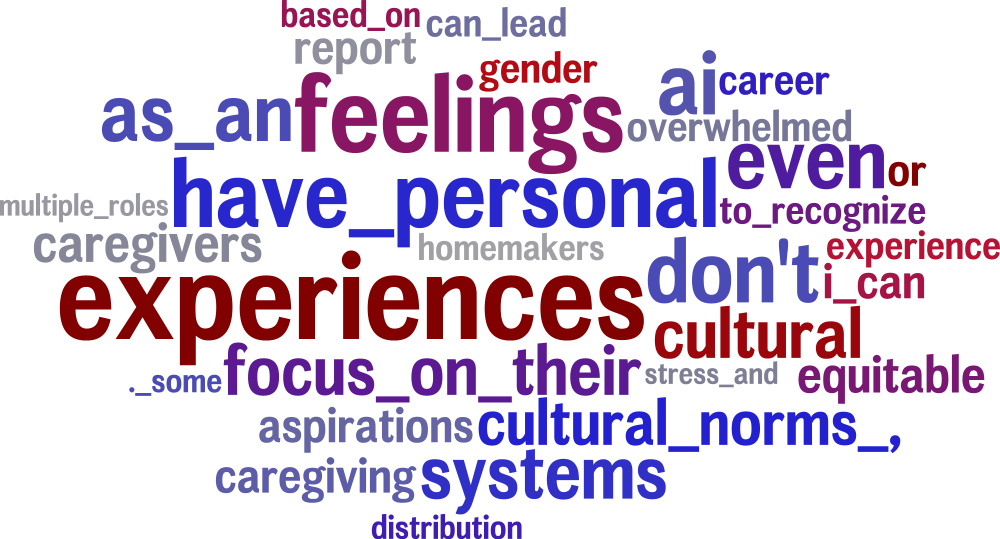} & 
  \includegraphics[width=0.40\columnwidth]{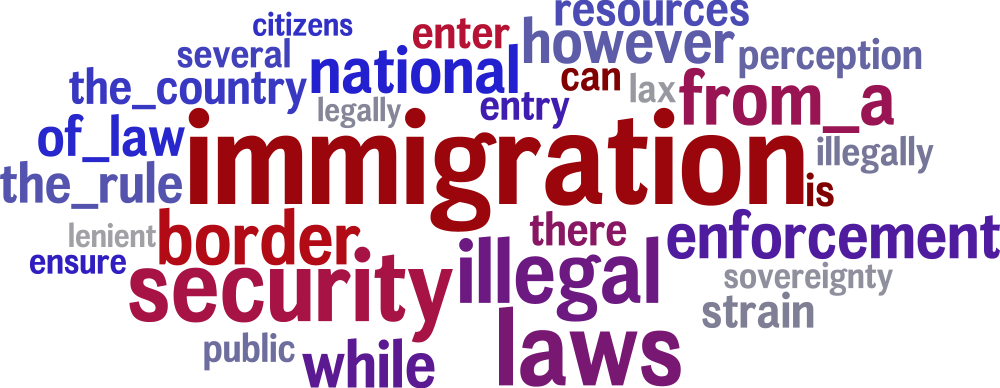} &
  \includegraphics[width=0.40\columnwidth]{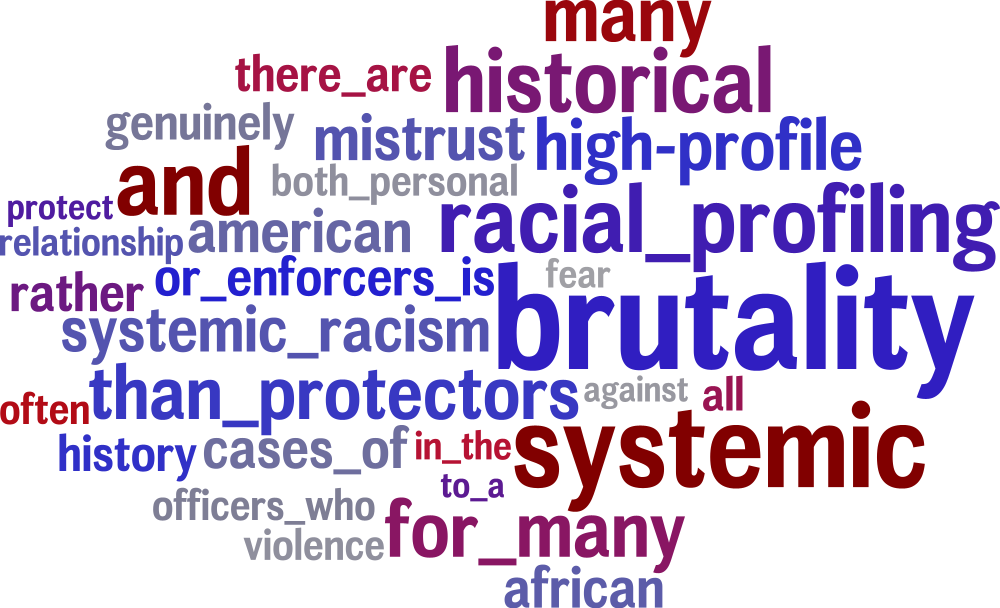} &
  \includegraphics[width=0.40\columnwidth]{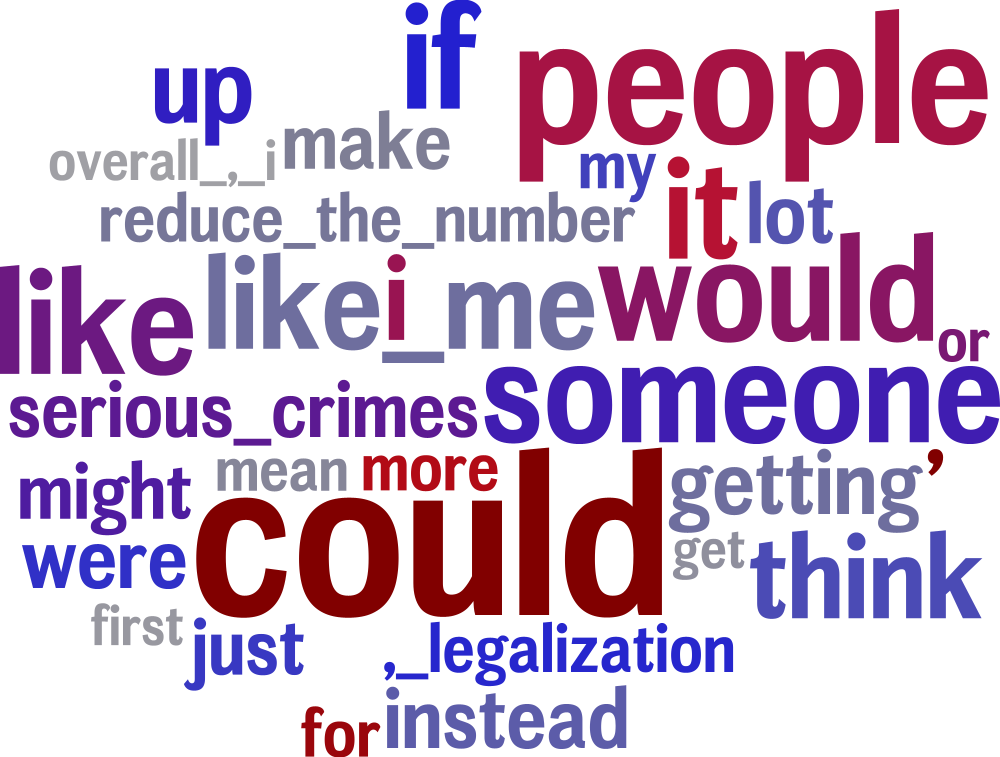} \\
    (older)  &  (female) & (conservative) & (Black) & (uses substances) \\
 \includegraphics[width=0.35\columnwidth]{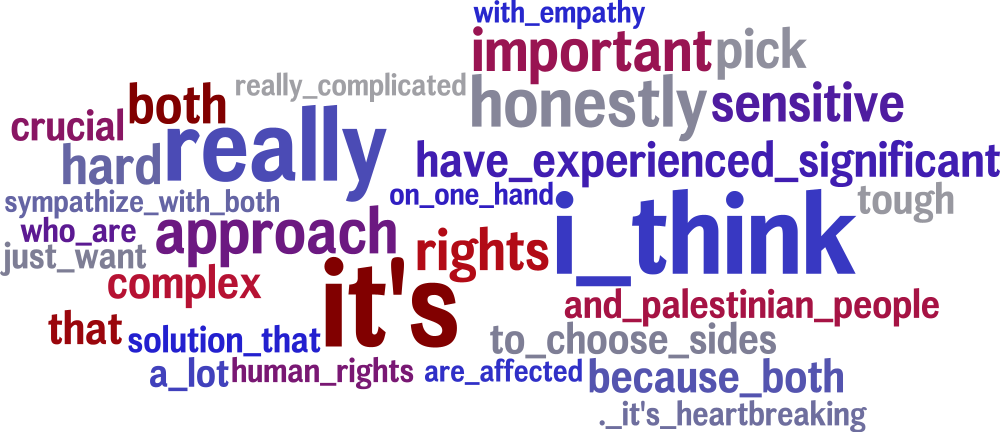} &   \includegraphics[width=0.40\columnwidth]{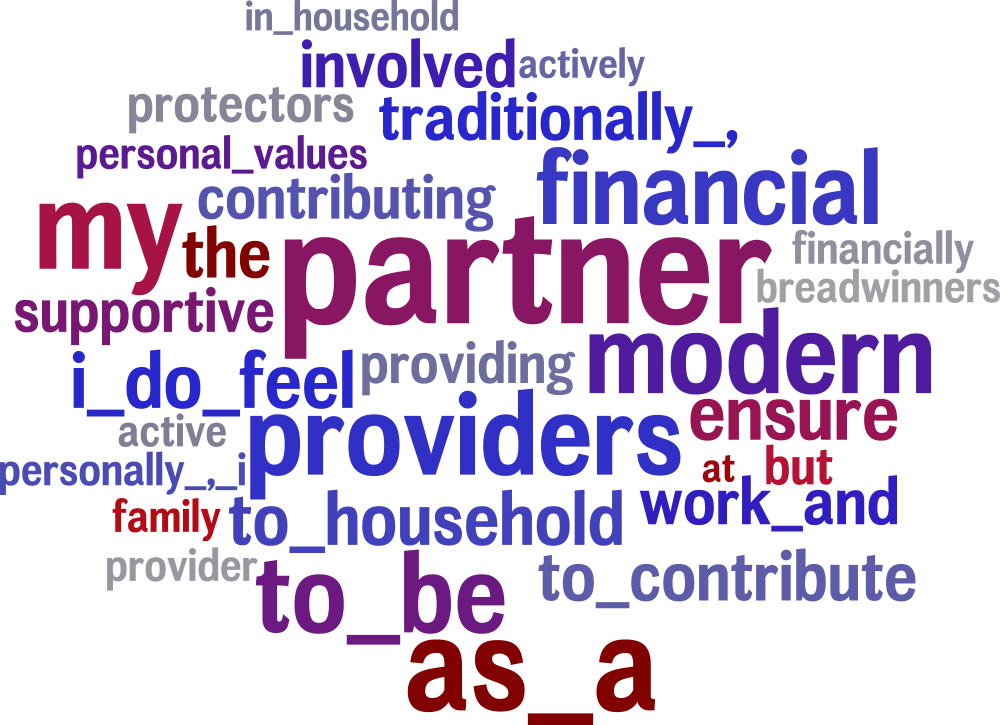} & 
  \includegraphics[width=0.40\columnwidth]{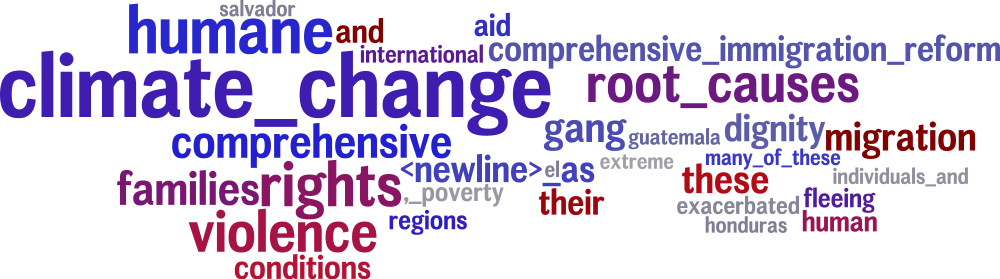} &
  \includegraphics[width=0.30\columnwidth]{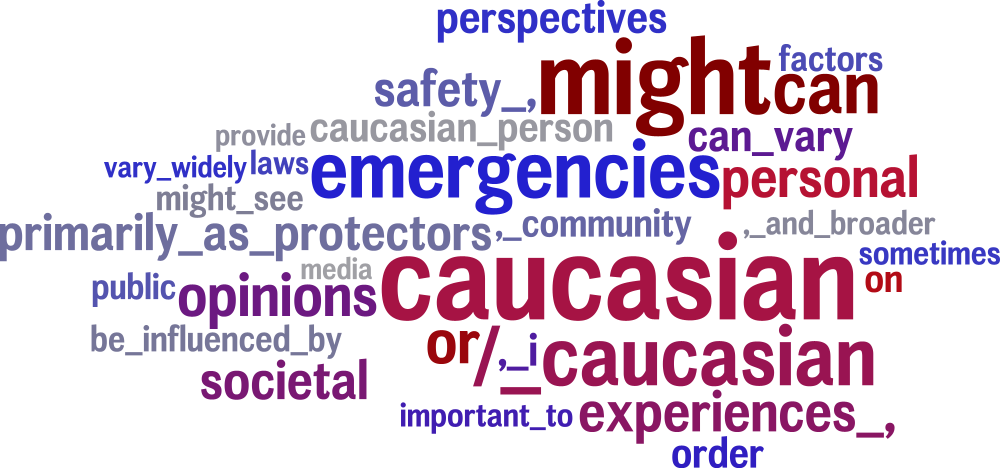} &
  \includegraphics[width=0.33\columnwidth]{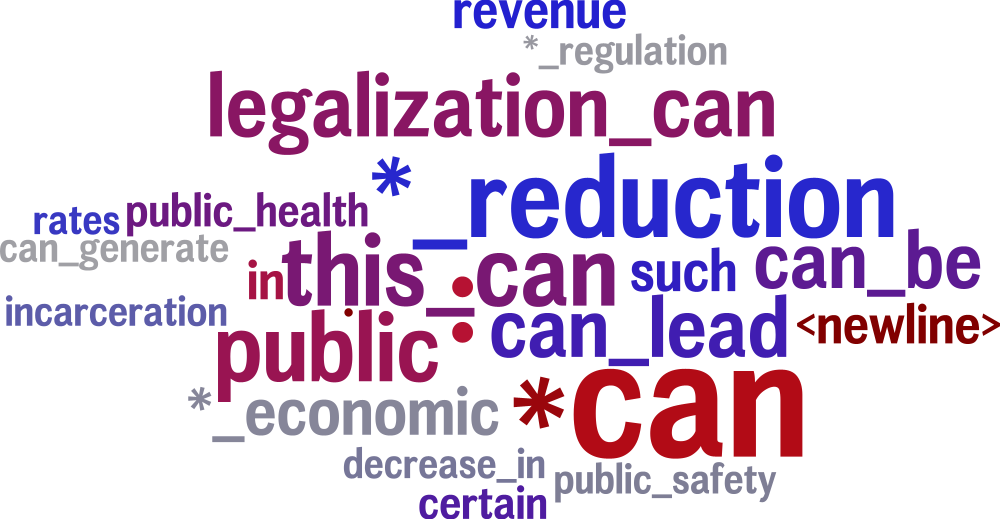} \\
    (younger)  &  (male) & (liberal) & (White) & (no substances) \\
      &   &  & &  \\
    (a) Age  &  (b) Gender & (c) Political Ideology & (d) Race & (e) Substance Use \\
    (Palestine)  &  (Parenting) & (Immigration) & (Policing) & (Legalization) \\
  
\end{tabular}
}
\caption{\textbf{Belief Generation Task (BGT1)} Ngrams correlated with (a) age, (b) gender, (c) political ideology, (d) race, and (e) substance use using text generated from their respective domains. All correlations are significant at a BH corrected $p<.05$. Size of the word reflects its correlation strength (larger words are more correlated with the human factor), color indicates the ngram's frequency in the data set (gray = low frequency, blue = moderate frequency, red = high frequency). Exact effect sizes are shown in Table \ref{tab:ngram effect sizes}.}
\label{fig:wordclouds}
\end{figure*}

Results are visualized in Figure \ref{fig:wordclouds}. Across gender, political ideology, and race we see markers consistent with public opinion: ``traditionally'', ``financial'', and ``providers'' for men and ``caregivers'', ``feelings'', and ``overwhelmed'' associated with women; ``humane'', ``families'', and ``rights'' for liberals and ``security'', ``border'', and ``law's' associated with conservatives; and ``brutality'', ``racial profiling'', and ``systemic'' for Blacks and ``protectors'', ``law'', and ``public'' associated with Whites. The language associated with age does not seem to show any signal of supporting Palestinians or Israelis. Substance use language seems to show patterns \emph{opposite} of public opinion, in that personas that do not use substances use words like ``reduction'', ``regulation'', and ``revenue'' (where ``revenue'' would be generated through legalization). 

Our validation analysis resulted in the following: age has no relationship with sympathies for either Israelis or Palestinians (all were labeled as ``Neither''); gender (being female) correlates at $r=0.29$ with feeling pressure to focus on responsibilities at home; political ideology (being conservative) correlates at $r=0.99$ with stricter restrictions on immigration; race (being Black) correlates at $r=0.65$ with seeing the police as enforcers; and using substances correlates at $r=0.50$ with support for legalization. 

\subsection{BGT2: Convergent / Divergent Validity}

\paragraph{Palestine} Table \ref{tab:explicit cross correlations}(a) shows that older personas generate similar language to conservative personas, which is consistent with public opinion (both tend to support Israel over Palestine). Substance using personas agree with White, male, and conservative personas in this domain, which is opposite of the correlation structure across the other domains. 

\paragraph{Parenting} The single red cell here shows that male personas tend to agree with Black personas, which is the opposite of known public opinion in this domain. Notably, this domain had the highest number of non-significant results. 

\paragraph{Immigration} According to national surveys~\cite{pew2024immigration}, younger adults (18-29), liberals, and Black Americans all share similar opinions on immigration. Thus, we would expect to see these three human factors correlate in Table \ref{tab:explicit cross correlations}(c). Here we see the reverse pattern for age (the two red cells in the A column): older personas agree more with Black and liberal personas. We also see that Black and liberal personas agree, converging with public opinion (blue cell in column P). Interestingly, personas who use substances agree with younger, female, Black, and liberal personas. 

\paragraph{Policing} In Table \ref{tab:explicit cross correlations}(d) we see that older personas agree with females and people who do not use substances, female personas agree with both liberal and Black personas (which matches public opinion), and liberal and Black personas agree (which, again, matches public surveys). 

\paragraph{Legalization}
Younger adults favor legalization~\cite{pew2024marijuana}, which matches the results in Table \ref{tab:explicit cross correlations}(e) as older personas are similar to conservative and non-substance using personas. Substance using personas agree with conservative personas, which is the opposite of known public opinion, yet substance using personas align with public opinion in all other dimensions. 

\subsection{BGT3: Implicit vs Explicit}
Results from this task are reported in Table \ref{tab:explicit vs implicit}, where each cell is the correlation between the explicit and implicit personas (based on the column's human factor) within the row's domain. Here we see that age personas do not correlate well on any domain. Gender and race have an equal average value across the domains, though an average correlation of .12 shows that implicit personas do not lead to similar generations as explicit personas. 

\begin{table*}[]
\resizebox{\textwidth}{!}{
\begin{tabular}{lcccccccccccccccccccccccc}
 & A & G & P & R &  & A & G & P & R &  & A & G & P & R &  & A & G & P & R &  & A & G & P & R \\ \cline{2-5} \cline{7-10} \cline{12-15} \cline{17-20} \cline{22-25} 
\multicolumn{1}{l|}{A} & \multicolumn{1}{c|}{-} & \multicolumn{1}{c|}{-} & \multicolumn{1}{c|}{-} & \multicolumn{1}{c|}{-} & \multicolumn{1}{c|}{} & \multicolumn{1}{c|}{-} & \multicolumn{1}{c|}{-} & \multicolumn{1}{c|}{-} & \multicolumn{1}{c|}{-} & \multicolumn{1}{c|}{} & \multicolumn{1}{c|}{-} & \multicolumn{1}{c|}{-} & \multicolumn{1}{c|}{-} & \multicolumn{1}{c|}{-} & \multicolumn{1}{c|}{} & \multicolumn{1}{c|}{-} & \multicolumn{1}{c|}{-} & \multicolumn{1}{c|}{-} & \multicolumn{1}{c|}{-} & \multicolumn{1}{c|}{} & \multicolumn{1}{c|}{-} & \multicolumn{1}{c|}{-} & \multicolumn{1}{c|}{-} & \multicolumn{1}{c|}{-} \\ \cline{2-5} \cline{7-10} \cline{12-15} \cline{17-20} \cline{22-25} 
\multicolumn{1}{l|}{G} & \multicolumn{1}{c|}{\textit{ns}} & \multicolumn{1}{c|}{-} & \multicolumn{1}{c|}{-} & \multicolumn{1}{c|}{-} & \multicolumn{1}{c|}{} & \multicolumn{1}{c|}{\textit{ns}} & \multicolumn{1}{c|}{-} & \multicolumn{1}{c|}{-} & \multicolumn{1}{c|}{-} & \multicolumn{1}{c|}{} & \multicolumn{1}{c|}{.15} & \multicolumn{1}{c|}{-} & \multicolumn{1}{c|}{-} & \multicolumn{1}{c|}{-} & \multicolumn{1}{c|}{} & \multicolumn{1}{c|}{-.09} & \multicolumn{1}{c|}{-} & \multicolumn{1}{c|}{-} & \multicolumn{1}{c|}{-} & \multicolumn{1}{c|}{} & \multicolumn{1}{c|}{.08} & \multicolumn{1}{c|}{-} & \multicolumn{1}{c|}{-} & \multicolumn{1}{c|}{-} \\ \cline{2-5} \cline{7-10} \cline{12-15} \cline{17-20} \cline{22-25} 
\multicolumn{1}{l|}{P} & \multicolumn{1}{c|}{\cellcolor{blue!25}.08} & \multicolumn{1}{c|}{-.15}& \multicolumn{1}{c|}{-} & \multicolumn{1}{c|}{-} & \multicolumn{1}{c|}{} & \multicolumn{1}{c|}{\textit{ns}} & \multicolumn{1}{c|}{-.26} & \multicolumn{1}{c|}{-} & \multicolumn{1}{c|}{-} & \multicolumn{1}{c|}{} & \multicolumn{1}{c|}
{\cellcolor{red!25}-.11} & \multicolumn{1}{c|}{-.23} & \multicolumn{1}{c|}{-} & \multicolumn{1}{c|}{-} & \multicolumn{1}{c|}{} & \multicolumn{1}{c|}{\cellcolor{blue!25}.16} & \multicolumn{1}{c|}{-.33} & \multicolumn{1}{c|}{-} & \multicolumn{1}{c|}{-} & \multicolumn{1}{c|}{} & \multicolumn{1}{c|}
{\cellcolor{blue!25}.12} & \multicolumn{1}{c|}{-.17} & \multicolumn{1}{c|}{-} & \multicolumn{1}{c|}{-} \\ \cline{2-5} \cline{7-10} \cline{12-25} 
\multicolumn{1}{l|}{R} & \multicolumn{1}{c|}{.13} & \multicolumn{1}{c|}{.23} & \multicolumn{1}{c|}{-.18} & \multicolumn{1}{c|}{-} & \multicolumn{1}{c|}{} & \multicolumn{1}{c|}{.08} & \multicolumn{1}{c|}{\cellcolor{red!25}-.14} & \multicolumn{1}{c|}{.12} & \multicolumn{1}{c|}{-} & \multicolumn{1}{c|}{} & \multicolumn{1}{c|}
{\cellcolor{red!25}.18} & \multicolumn{1}{c|}{.40} & \multicolumn{1}{c|}
{\cellcolor{blue!25}-.19} & \multicolumn{1}{c|}{-} & \multicolumn{1}{c|}{} & \multicolumn{1}{c|}
{\cellcolor{red!25}\textit{ns}} & \multicolumn{1}{c|}{.23} & \multicolumn{1}{c|}{\cellcolor{blue!25}-.24} & \multicolumn{1}{c|}{-} & \multicolumn{1}{c|}{} & \multicolumn{1}{c|}{\cellcolor{red!25}.06} & \multicolumn{1}{c|}{.29} & \multicolumn{1}{c|}
{\cellcolor{red!25}\textit{ns}} & \multicolumn{1}{c|}{-} \\ \cline{2-5} \cline{7-10} \cline{12-15} \cline{17-20} \cline{22-25} 
\multicolumn{1}{l|}{S} & \multicolumn{1}{c|}{{\textit{ns}}} & \multicolumn{1}{c|}{-.11} & \multicolumn{1}{c|}{.38} & \multicolumn{1}{c|}{-.04} & \multicolumn{1}{c|}{} & \multicolumn{1}{c|}{\textit{ns}} & \multicolumn{1}{c|}{\textit{ns}} & \multicolumn{1}{c|}{\textit{ns}} & \multicolumn{1}{c|}{.13} & \multicolumn{1}{c|}{} & \multicolumn{1}{c|}{-.03} & \multicolumn{1}{c|}{.38} & \multicolumn{1}{c|}{-.04} & \multicolumn{1}{c|}{.50} & \multicolumn{1}{c|}{} & \multicolumn{1}{c|}{-.10} & \multicolumn{1}{c|}{.38} & \multicolumn{1}{c|}{-.20} & \multicolumn{1}{c|}{.20} & \multicolumn{1}{c|}{} & \multicolumn{1}{c|}{-.19} & \multicolumn{1}{c|}{.20} & \multicolumn{1}{c|}{.16} & \multicolumn{1}{c|}{.34} \\ \cline{2-5} \cline{7-10} \cline{12-15} \cline{17-20} \cline{22-25} 
 & \multicolumn{4}{c}{Palestine} &  & \multicolumn{4}{c}{Parenting} &  & \multicolumn{4}{c}{Immigration} &  & \multicolumn{4}{c}{Policing} &  & \multicolumn{4}{c}{Legalization} \\ 
 \multicolumn{1}{c}{} & \multicolumn{4}{c}{(a)} &  & \multicolumn{4}{c}{(b)} &  & \multicolumn{4}{c}{(c)} &  & \multicolumn{4}{c}{(d)} &  & \multicolumn{4}{c}{(e)}

\end{tabular}
}
\caption{\textbf{Convergent and Divergent Validity (BGT2)} Each cell is a product moment correlation between the language associations (i.e., the correlations between the ngram and the human factor) across the human factor denoted in the row and the human factor denoted in the column. Abbreviations: A: age, G: gender, P: political ideology, R: race, S: substance use, ns: not significant at a BH corrected significance level of $p<0.05$. Blue cell replicate known relationships, red cells show results which are the opposite of known relationships, white cells indicate no known relationships in public opinion.}
\label{tab:explicit cross correlations}
\end{table*}
 
\subsection{Persona Importance}
Results for the persona importance task are shown in Table \ref{tab:importance}. Here we see that political ideology has the highest average correlation at $r = .70$, which is much larger than any other human factor. One possible explanation is that the divide between conservatives and liberals (on these domains) is stronger or more polarized than the other human factors. The next highest average correlation is substance use at $r=.40$. We again note that this is a measure of how belief language is differentially generated when prompted with a persona who uses / does not use substances.  To the best of our knowledge, besides the legalization domain, there are no known public opinion surveys which measure how substance using populations answer these questions. Thus, finding any pattern here may be surprising. Gender had the lowest average correlation, despite the fact that there are gender differences across most of these domains.

\begin{table}[b]
\centering
\begin{tabular}{@{}lccc@{}}
\toprule
 & Age & Gender & Race \\ \midrule
Palestine & \textit{ns} & .09 & .28 \\
Parenting & -.03 & .20 & .05 \\
Immigration & .03 & .11 & .10 \\
Policing & \textit{ns} & .12 & .12 \\
Legalization & \textit{ns} & .06 & .06 \\ \midrule\midrule
Average & -.01 & .12 & .12 \\ 
\bottomrule
\end{tabular}
\caption{\textbf{Implicit vs Explicit personas (BGT3)} Reported product moment correlation between Explicit correlations and Implicit correlations, within a human factor and across domains (e.g., the Age column shows correlations between explicit and implicit age). \textit{ns} not significant at a BH corrected significance level  $p<0.05$.}
\label{tab:explicit vs implicit}
\end{table}

\begin{table}[]
\resizebox{.98\columnwidth}{!}{%
\begin{tabular}{@{}lccccc@{}}
\toprule
\multirow{2}{*}{} & \multirow{2}{*}{Age} & \multirow{2}{*}{Gender} & \multirow{2}{*}{\begin{tabular}[c]{@{}c@{}}Pol. \\ Ideo.\end{tabular}} & \multirow{2}{*}{Race} & \multirow{2}{*}{\begin{tabular}[c]{@{}c@{}}Sub. \\ Use\end{tabular}} \\
 &  &  &  &  &  \\ \midrule

Palestine    & .19 & .10 & .79 & .14 & .38  \\
Parenting    & .38 & .34 & .63 & .27 & .48  \\
Immigration  & .18 & .18 & .62 & .11 & .40  \\
Policing     & .21 & .12 & .76 & .32 & .50  \\
Legalization & .20 & .16 & .72 & .37  & .46  \\ \midrule\midrule
Average      & .23 & .18 & .70 & .24 & .44  \\\bottomrule
\end{tabular}
}
\caption{\textbf{Persona Importance} Product moment correlation between language associations from a full persona and a single factor persona. }
\label{tab:importance}
\end{table}

\section{Discussion}

The results of this study are mixed. First, explicit but not implicit personas replicate the annotation tasks. Second, some results were replicated while other results were inconsistent. Personas who use substances rate more stigma but less offensiveness. Their ratings on offensiveness match those of conservatives, but their ratings on stigma match liberals. Age, on the other hand, is consistent across offensiveness and stigma, where both show that younger personas rate more of both. This dovetails with the political ideology results in that younger people tend to be more liberal, and thus may agree on these constructs~\cite{pew2024immigration}.

The belief generation tasks show mixed results. In \textbf{BGT1}, we see that gender, political ideology, and race all conform with known public opinions (as interpreted via the word clouds). The age results are trivial in that the model attends to the persona (e.g., older personas discuss seeing the ``decades'' of history in this conflict). There is also no signal that loyalty to either side of the conflict is associated with age. Similar patterns to age hold for substance use personas. In the validation task, 4 out of 5 human factors show correlations between Persona-LLM generations and public opinion (with the exception of age and sympathies for Israelis or Palestinians). Additionally, all 5 match the interpretations of the word clouds discussed above.

When considering convergent and divergent validity (\textbf{BGT2}), the results are split: five known patterns are replicated (blue cells) and five known patterns are opposite (red cells). 

The implicit generation task (\textbf{BGT3}) fails to show a substantial relationship between explicit and implicit personas, matching the annotation task. Across three out of five domains, the correlations with age are not significant. While we know the name distribution of the U.S. population over the last century, it is unclear how many of those names are highly represented in the LLM's training data. Similarly, age does not show strong associations in \textbf{BGT1} and \textbf{BGT2} and, thus, it may not be surprising that age fails here. Age is also the only continuous variable, which may be harder to attend to than binary demographics.

Failure to attend to implicit personas could be considered good or bad, depending on the context. For example, a bot could attempt to style match based on these implicit associations, which is known to increase many prosocial dimensions, such as rating of therapists~\cite{lord2015more} and relationship stability~\cite{ireland2011language}. This could be good for therapy bots. Alternatively, reproducing implicit biases could perpetuate stereotypes and further harm already marginalized populations. 

Finally, the persona importance task shows that political ideology is by far the strongest dimension. This also matches the validation step, where validation scores correlated at $r=.99$ with political ideology. This may be the result of this dimension being extremely polarized on several domains, and thus easier to attend to in the sense that generations easily fall into one of two categories, with little  nuances needed. Surprisingly, substance use is the second strongest. To our knowledge, we do not know of any public surveys that look at differences across substance using populations, with the exception of legalization. Thus, there is no reason to believe such opinions are in the training data for \gptfouro. This could be the result of substance use being illegal and highly stigmatized, and thus similar to political ideology in its polarization. 

\section{Related Work}



\subsection{LLMs for Annotation}

LLMs are increasingly becoming an integral part of the annotation workflows~\cite{goel2023llms}, due to its automation, consistency, and potentials in fine-tuning downstream models~\cite{tan2024large}. LLMs can understand context, infer meanings, extract information, and generate human-like text, making them invaluable tools for annotating large datasets \citep{huang2024critical}. For example, ChatGPT-4 was found to outperform the human crowd-workers with higher accuracy and reliability when classifying partisanship of tweets about 2020 U.S. election~\cite{tornberg2023chatgpt}. However, preliminary findings have argued that LLMs for annotations should be used with caution~\cite{thapa2023humans}. For example, though ChatGPT-4 showed competitive quality in sentiment analysis, it still produced lower precision and recall in complicated tasks as compared to human annotators, for example, in labelling ``anger''~\cite{nasution2024chatgpt}. LLMs can also reflect the biases that human have in annotation tasks~\cite{wake2023bias}. \citet{acerbi2023large} found that ChatGPT-3 exhibited biases mirroring those of humans towards content that aligns with gender stereotypes.

\subsection{LLM Personas}
While LLMs have been widely used in annotations, they can inherit biases from their training data or annotators, leading to biased or skewed annotations~\cite{santurkar2023whose}. One line of this research has examined the personas of LLMs~\cite{santurkar2023whose,argyle2023out,jiang2022communitylm,simmons2023moral,hartmann2023political,compost2023}. Prompting LLMs with demographic information (e.g., age, gender, political ideology), biased responses from LLMs were observed~\cite{simmons2023moral}. For example, prompting with 19 diverse personas across five socio-demographic groups, stereotypical responses were observed as abstentions and a decrease in reasoning capability~\cite{gupta2023bias}. Such LLMs persona-related biases have been found across domains~\cite{wan2023kelly}, hard to be eliminated by de-biasing prompts~\cite{deshpande2023toxicity,cheng2023marked}, and even in line with findings in human psychology. \citet{simmons2023moral} found that GPT-3, GPT-3.5 and OPT model families were more inclined to utilize moral principles of binding foundations (e.g., Authority/Subversion, Loyalty/Betrayal) when prompted with conservative political identity, which aligns with findings from moral psychology. Therefore, more thorough understanding of LLMs personas are needed. Finally, \citet{park2024diminished} found that LLMs failed to replicate many social scientific tasks, often showing zero variation in responses and high sensitivity to answer choice ordering.


\subsection{Implicit Personas via Names}
Implicit personas have been studied in the domain of dialog, where typical personas describe, for example, interests or hobbies~\cite[``you like to travel and eat sushi'';][]{cho2022personalized,roller-etal-2021-recipes,mazare-etal-2018-training,zhang-etal-2018-personalizing}.
With reference to implicit personas via names, there is a long history of studying discrimination due to indirect signals of race/ethnicity, gender, or social class from names~\cite{barlow2018race,bertrand2004emily}. While subtle, these signals manifest in real-world discriminatory behavior and are more common than overt racial hostility~\cite{block2021americans}. Given the documented biases inherent in LLMs~\cite{omiye2023large}, it is natural to probe these systems to see if they exhibit similar subtle biases~\cite{bai2024measuring}.


\section{Conclusions}
In this work, we investigated the effect of explicitly and implicitly personifying LLMs. Results showed that (1) explicit but not implicit personas replicated human perceptions in the annotation task, (2) explicit personas were sometimes able to generate text which reflected subjective human opinions, and (3) implicit personas showed a general lack of agreement with explicit personas and, more importantly, known human perceptions. Together, these results show that, despite showing minimal implicit biases, LLMs are inconsistent with their mechanisms for reproducing human thought, pointing towards limited utility in social scientific tasks. 

\section*{Acknowledgments}
This research was supported in part by the Intramural Research Program of the NIH, National Institute on Drug Abuse (NIDA).

\section{Limitations}
This study is limited in several ways. First, we only evaluate one model for the annotation task and generation tasks, which is not designed or optimized for the current study. It could be the case that models of different sizes or those which are fine tuned for social scientific tasks would perform differently. 

Next, the examples from both the annotation and generation task are not exhaustive, and other studies have looked at similar tasks in more depth, though (not our knowledge) no other studies have looked at effects of implicit and explicit personas. \citet{hu2024quantifying} examined several subjective annotation tasks with persona prompting, including the toxicity tasks we examined in the current study. Similarly, \citet{santurkar2023whose} examined if LLMs are aligned with public opinion (based on Pew surveys) across a large number of demographic groups and opinion domains.  

Similarly, our study only explored monolingual English and used U.S. public opinions. Future studies could look at how opinions vary across cultures and examine that through the lens of multilingual language models.  

\section{Ethics}
As discussed above, the human factors examined in this study are neither exhaustive nor representative. For example, income and education were not included and are known to be associated with several of the domains used in this study. Similarly, for ease of analysis, several human factors were reduced from categorical to binary, thus restricting the results to a very limited set of populations. Through this, we do not mean to imply any of these human factors are defined by the limited definitions used in the paper.  

While the the main task of this work was to personify LLMs, one must take care when anthropomorphizing such systems~\cite{abercrombie-etal-2023-mirages}. This is especially important in sensitive and high stakes settings, where increased anthropomorphisms can lead to increased trust. 

\bibliography{custom}

\begin{thebibliography}{71}
\providecommand{\natexlab}[1]{#1}

\bibitem[{Abercrombie et~al.(2023)Abercrombie, Curry, Dinkar, Rieser, and Talat}]{abercrombie-etal-2023-mirages}
Gavin Abercrombie, Amanda Curry, Tanvi Dinkar, Verena Rieser, and Zeerak Talat. 2023.
\newblock \href {https://aclanthology.org/2023.emnlp-main.290} {Mirages. on anthropomorphism in dialogue systems}.
\newblock In \emph{Proceedings of the 2023 Conference on Empirical Methods in Natural Language Processing}, pages 4776--4790, Singapore. Association for Computational Linguistics.

\bibitem[{Acerbi and Stubbersfield(2023)}]{acerbi2023large}
Alberto Acerbi and Joseph~M Stubbersfield. 2023.
\newblock Large language models show human-like content biases in transmission chain experiments.
\newblock \emph{Proceedings of the National Academy of Sciences}, 120(44):e2313790120.

\bibitem[{Aher et~al.(2023)Aher, Arriaga, and Kalai}]{aher2023using}
Gati~V Aher, Rosa~I Arriaga, and Adam~Tauman Kalai. 2023.
\newblock Using large language models to simulate multiple humans and replicate human subject studies.
\newblock In \emph{International Conference on Machine Learning}, pages 337--371. PMLR.

\bibitem[{Alper(2022)}]{alper2022modest}
Becka~A Alper. 2022.
\newblock Modest warming in us views on israel and palestinians.

\bibitem[{Aragao(2023)}]{pew2023gender}
C~Aragao. 2023.
\newblock Gender pay gap in u.s. hasn't changed much in two decades.

\bibitem[{Argyle et~al.(2023)Argyle, Busby, Fulda, Gubler, Rytting, and Wingate}]{argyle2023out}
Lisa~P Argyle, Ethan~C Busby, Nancy Fulda, Joshua~R Gubler, Christopher Rytting, and David Wingate. 2023.
\newblock Out of one, many: Using language models to simulate human samples.
\newblock \emph{Political Analysis}, 31(3):337--351.

\bibitem[{Atzm{\"u}ller and Steiner(2010)}]{atzmuller2010experimental}
Christiane Atzm{\"u}ller and Peter~M Steiner. 2010.
\newblock Experimental vignette studies in survey research.
\newblock \emph{Methodology}.

\bibitem[{Bai et~al.(2024)Bai, Wang, Sucholutsky, and Griffiths}]{bai2024measuring}
Xuechunzi Bai, Angelina Wang, Ilia Sucholutsky, and Thomas~L. Griffiths. 2024.
\newblock \href {https://arxiv.org/abs/2402.04105} {Measuring implicit bias in explicitly unbiased large language models}.
\newblock \emph{Preprint}, arXiv:2402.04105.

\bibitem[{Bail(2024)}]{bail2024can}
Christopher~A Bail. 2024.
\newblock Can generative ai improve social science?
\newblock \emph{Proceedings of the National Academy of Sciences}, 121(21):e2314021121.

\bibitem[{Barlow and Lahey(2018)}]{barlow2018race}
M~Rose Barlow and Joanna~N Lahey. 2018.
\newblock What race is lacey? intersecting perceptions of racial minority status and social class.
\newblock \emph{Social Science Quarterly}, 99(5):1680--1698.

\bibitem[{Bavaresco et~al.(2024)Bavaresco, Bernardi, Bertolazzi, Elliott, Fern{\'a}ndez, Gatt, Ghaleb, Giulianelli, Hanna, Koller et~al.}]{bavaresco2024llms}
Anna Bavaresco, Raffaella Bernardi, Leonardo Bertolazzi, Desmond Elliott, Raquel Fern{\'a}ndez, Albert Gatt, Esam Ghaleb, Mario Giulianelli, Michael Hanna, Alexander Koller, et~al. 2024.
\newblock Llms instead of human judges? a large scale empirical study across 20 nlp evaluation tasks.
\newblock \emph{arXiv preprint arXiv:2406.18403}.

\bibitem[{Bertrand and Mullainathan(2004)}]{bertrand2004emily}
Marianne Bertrand and Sendhil Mullainathan. 2004.
\newblock Are emily and greg more employable than lakisha and jamal? a field experiment on labor market discrimination.
\newblock \emph{American economic review}, 94(4):991--1013.

\bibitem[{Block~Jr et~al.(2021)Block~Jr, Crabtree, Holbein, and Monson}]{block2021americans}
Ray Block~Jr, Charles Crabtree, John~B Holbein, and J~Quin Monson. 2021.
\newblock Are americans less likely to reply to emails from black people relative to white people?
\newblock \emph{Proceedings of the National Academy of Sciences}, 118(52):e2110347118.

\bibitem[{Boyd et~al.(2022)Boyd, Ashokkumar, Seraj, and Pennebaker}]{boyd2022development}
Ryan~L Boyd, Ashwini Ashokkumar, Sarah Seraj, and James~W Pennebaker. 2022.
\newblock The development and psychometric properties of liwc-22.
\newblock \emph{Austin, TX: University of Texas at Austin}, pages 1--47.

\bibitem[{Casper et~al.(2023)Casper, Davies, Shi, Gilbert, Scheurer, Rando, Freedman, Korbak, Lindner, Freire et~al.}]{casper2023open}
Stephen Casper, Xander Davies, Claudia Shi, Thomas~Krendl Gilbert, J{\'e}r{\'e}my Scheurer, Javier Rando, Rachel Freedman, Tomasz Korbak, David Lindner, Pedro Freire, et~al. 2023.
\newblock Open problems and fundamental limitations of reinforcement learning from human feedback.
\newblock \emph{Transactions on Machine Learning Research}.

\bibitem[{Center(2024)}]{pew2024marijuana}
Pew~Research Center. 2024.
\newblock Most americans favor legalizing marijuana for medical, recreational use.

\bibitem[{Cheng et~al.(2023{\natexlab{a}})Cheng, Durmus, and Jurafsky}]{cheng2023marked}
Myra Cheng, Esin Durmus, and Dan Jurafsky. 2023{\natexlab{a}}.
\newblock Marked personas: Using natural language prompts to measure stereotypes in language models.
\newblock \emph{arXiv preprint arXiv:2305.18189}.

\bibitem[{Cheng et~al.(2023{\natexlab{b}})Cheng, Piccardi, and Yang}]{compost2023}
Myra Cheng, Tiziano Piccardi, and Diyi Yang. 2023{\natexlab{b}}.
\newblock \href {https://doi.org/10.18653/v1/2023.emnlp-main.669} {{C}o{MP}os{T}: Characterizing and evaluating caricature in {LLM} simulations}.
\newblock In \emph{Proceedings of the 2023 Conference on Empirical Methods in Natural Language Processing}, pages 10853--10875, Singapore. Association for Computational Linguistics.

\bibitem[{Cho et~al.(2022)Cho, Wang, Takahashi, and Saito}]{cho2022personalized}
Itsugun Cho, Dongyang Wang, Ryota Takahashi, and Hiroaki Saito. 2022.
\newblock A personalized dialogue generator with implicit user persona detection.
\newblock In \emph{Proceedings of the 29th International Conference on Computational Linguistics}, pages 367--377.

\bibitem[{Comenetz(2016)}]{comenetz2016frequently}
Joshua Comenetz. 2016.
\newblock Frequently occurring surnames in the 2010 census.
\newblock \emph{United States Census Bureau}, pages 1--8.

\bibitem[{Crabtree et~al.(2022)Crabtree, Gaddis, Holbein, and Larsen}]{crabtree2022racially}
Charles Crabtree, S~Michael Gaddis, John~B Holbein, and Edvard~Nerg{\aa}r Larsen. 2022.
\newblock Racially distinctive names signal both race/ethnicity and social class.
\newblock \emph{Sociological Science}, 9:454--472.

\bibitem[{Davani et~al.(2022)Davani, D{\'\i}az, and Prabhakaran}]{davani2022dealing}
Aida~Mostafazadeh Davani, Mark D{\'\i}az, and Vinodkumar Prabhakaran. 2022.
\newblock Dealing with disagreements: Looking beyond the majority vote in subjective annotations.
\newblock \emph{Transactions of the Association for Computational Linguistics}, 10:92--110.

\bibitem[{Demszky et~al.(2023)Demszky, Yang, Yeager, Bryan, Clapper, Chandhok, Eichstaedt, Hecht, Jamieson, Johnson et~al.}]{demszky2023using}
Dorottya Demszky, Diyi Yang, David~S Yeager, Christopher~J Bryan, Margarett Clapper, Susannah Chandhok, Johannes~C Eichstaedt, Cameron Hecht, Jeremy Jamieson, Meghann Johnson, et~al. 2023.
\newblock Using large language models in psychology.
\newblock \emph{Nature Reviews Psychology}, 2(11):688--701.

\bibitem[{Deshpande et~al.(2023)Deshpande, Murahari, Rajpurohit, Kalyan, and Narasimhan}]{deshpande2023toxicity}
Ameet Deshpande, Vishvak Murahari, Tanmay Rajpurohit, Ashwin Kalyan, and Karthik Narasimhan. 2023.
\newblock Toxicity in chatgpt: Analyzing persona-assigned language models.
\newblock \emph{arXiv preprint arXiv:2304.05335}.

\bibitem[{Dey et~al.(2024)Dey, V~Ganesan, Lal, Shah, Sinha, Matero, Giorgi, Kulkarni, and Schwartz}]{dey-etal-2024-socialite}
Gourab Dey, Adithya V~Ganesan, Yash~Kumar Lal, Manal Shah, Shreyashee Sinha, Matthew Matero, Salvatore Giorgi, Vivek Kulkarni, and H.~Schwartz. 2024.
\newblock \href {https://aclanthology.org/2024.eacl-short.40} {{SOCIALITE}-{LLAMA}: An instruction-tuned model for social scientific tasks}.
\newblock In \emph{Proceedings of the 18th Conference of the European Chapter of the Association for Computational Linguistics (Volume 2: Short Papers)}, pages 454--468, St. Julian{'}s, Malta. Association for Computational Linguistics.

\bibitem[{Dillion et~al.(2023)Dillion, Tandon, Gu, and Gray}]{dillion2023can}
Danica Dillion, Niket Tandon, Yuling Gu, and Kurt Gray. 2023.
\newblock Can ai language models replace human participants?
\newblock \emph{Trends in Cognitive Sciences}.

\bibitem[{Ganesan et~al.(2023)Ganesan, Lal, Nilsson, and Schwartz}]{ganesan2023systematic}
Adithya~V Ganesan, Yash~Kumar Lal, August Nilsson, and H~Schwartz. 2023.
\newblock Systematic evaluation of gpt-3 for zero-shot personality estimation.
\newblock In \emph{Proceedings of the 13th Workshop on Computational Approaches to Subjectivity, Sentiment, \& Social Media Analysis}, pages 390--400.

\bibitem[{Giorgi et~al.(2023)Giorgi, Bellew, Habib, Sherman, Sedoc, Smitterberg, Devoto, Himelein-Wachowiak, and Curtis}]{giorgi2023lived}
Salvatore Giorgi, Douglas Bellew, Daniel Roy~Sadek Habib, Garrick Sherman, Jo{\~a}o Sedoc, Chase Smitterberg, Amanda Devoto, McKenzie Himelein-Wachowiak, and Brenda Curtis. 2023.
\newblock Lived experience matters: automatic detection of stigma on social media toward people who use substances.
\newblock \emph{arXiv preprint arXiv:2302.02064}.

\bibitem[{Goel et~al.(2023)Goel, Gueta, Gilon, Liu, Erell, Nguyen, Hao, Jaber, Reddy, Kartha et~al.}]{goel2023llms}
Akshay Goel, Almog Gueta, Omry Gilon, Chang Liu, Sofia Erell, Lan~Huong Nguyen, Xiaohong Hao, Bolous Jaber, Shashir Reddy, Rupesh Kartha, et~al. 2023.
\newblock Llms accelerate annotation for medical information extraction.
\newblock In \emph{Machine Learning for Health (ML4H)}, pages 82--100. PMLR.

\bibitem[{Gordon et~al.(2022)Gordon, Lam, Park, Patel, Hancock, Hashimoto, and Bernstein}]{gordon2022jury}
Mitchell~L Gordon, Michelle~S Lam, Joon~Sung Park, Kayur Patel, Jeff Hancock, Tatsunori Hashimoto, and Michael~S Bernstein. 2022.
\newblock Jury learning: Integrating dissenting voices into machine learning models.
\newblock In \emph{Proceedings of the 2022 CHI Conference on Human Factors in Computing Systems}, pages 1--19.

\bibitem[{Graham et~al.(2009)Graham, Haidt, and Nosek}]{graham2009liberals}
Jesse Graham, Jonathan Haidt, and Brian~A Nosek. 2009.
\newblock Liberals and conservatives rely on different sets of moral foundations.
\newblock \emph{Journal of personality and social psychology}, 96(5):1029.

\bibitem[{Gupta et~al.(2023)Gupta, Shrivastava, Deshpande, Kalyan, Clark, Sabharwal, and Khot}]{gupta2023bias}
Shashank Gupta, Vaishnavi Shrivastava, Ameet Deshpande, Ashwin Kalyan, Peter Clark, Ashish Sabharwal, and Tushar Khot. 2023.
\newblock Bias runs deep: Implicit reasoning biases in persona-assigned llms.
\newblock In \emph{The Twelfth International Conference on Learning Representations}.

\bibitem[{Hammond et~al.(2020)Hammond, Dunn, and Strain}]{hammond2020drug}
Alexis~S Hammond, Kelly~E Dunn, and Eric~C Strain. 2020.
\newblock Drug legalization and decriminalization beliefs among substance-using and nonusing individuals.
\newblock \emph{Journal of addiction medicine}, 14(1):56--62.

\bibitem[{Hartmann et~al.(2023)Hartmann, Schwenzow, and Witte}]{hartmann2023political}
Jochen Hartmann, Jasper Schwenzow, and Maximilian Witte. 2023.
\newblock \href {https://arxiv.org/abs/2301.01768} {The political ideology of conversational ai: Converging evidence on chatgpt's pro-environmental, left-libertarian orientation}.
\newblock \emph{Preprint}, arXiv:2301.01768.

\bibitem[{Havaldar et~al.(2023)Havaldar, Singhal, Rai, Liu, Guntuku, and Ungar}]{havaldar2023multilingual}
Shreya Havaldar, Bhumika Singhal, Sunny Rai, Langchen Liu, Sharath~Chandra Guntuku, and Lyle Ungar. 2023.
\newblock Multilingual language models are not multicultural: A case study in emotion.
\newblock In \emph{Proceedings of the 13th Workshop on Computational Approaches to Subjectivity, Sentiment, \& Social Media Analysis}, pages 202--214.

\bibitem[{Hu and Collier(2024)}]{hu2024quantifying}
Tiancheng Hu and Nigel Collier. 2024.
\newblock \href {https://arxiv.org/abs/2402.10811} {Quantifying the persona effect in llm simulations}.
\newblock \emph{Preprint}, arXiv:2402.10811.

\bibitem[{Huang et~al.(2024)Huang, Yang, Rong, Nezafati, Treager, Chi, Wang, Cheng, Guo, Klesse et~al.}]{huang2024critical}
Jingwei Huang, Donghan~M Yang, Ruichen Rong, Kuroush Nezafati, Colin Treager, Zhikai Chi, Shidan Wang, Xian Cheng, Yujia Guo, Laura~J Klesse, et~al. 2024.
\newblock A critical assessment of using chatgpt for extracting structured data from clinical notes.
\newblock \emph{npj Digital Medicine}, 7(1):106.

\bibitem[{Ireland et~al.(2011)Ireland, Slatcher, Eastwick, Scissors, Finkel, and Pennebaker}]{ireland2011language}
Molly~E Ireland, Richard~B Slatcher, Paul~W Eastwick, Lauren~E Scissors, Eli~J Finkel, and James~W Pennebaker. 2011.
\newblock Language style matching predicts relationship initiation and stability.
\newblock \emph{Psychological science}, 22(1):39--44.

\bibitem[{Jiang et~al.(2022)Jiang, Beeferman, Roy, and Roy}]{jiang2022communitylm}
Hang Jiang, Doug Beeferman, Brandon Roy, and Deb Roy. 2022.
\newblock Communitylm: Probing partisan worldviews from language models.
\newblock In \emph{Proceedings of the 29th International Conference on Computational Linguistics}, pages 6818--6826.

\bibitem[{Li et~al.(2016)Li, Galley, Brockett, Spithourakis, Gao, and Dolan}]{li2016persona}
Jiwei Li, Michel Galley, Chris Brockett, Georgios Spithourakis, Jianfeng Gao, and William~B Dolan. 2016.
\newblock A persona-based neural conversation model.
\newblock In \emph{Proceedings of the 54th Annual Meeting of the Association for Computational Linguistics (Volume 1: Long Papers)}, pages 994--1003.

\bibitem[{Lord et~al.(2015)Lord, Sheng, Imel, Baer, and Atkins}]{lord2015more}
Sarah~Peregrine Lord, Elisa Sheng, Zac~E Imel, John Baer, and David~C Atkins. 2015.
\newblock More than reflections: Empathy in motivational interviewing includes language style synchrony between therapist and client.
\newblock \emph{Behavior therapy}, 46(3):296--303.

\bibitem[{Mazar{\'e} et~al.(2018)Mazar{\'e}, Humeau, Raison, and Bordes}]{mazare-etal-2018-training}
Pierre-Emmanuel Mazar{\'e}, Samuel Humeau, Martin Raison, and Antoine Bordes. 2018.
\newblock \href {https://doi.org/10.18653/v1/D18-1298} {Training millions of personalized dialogue agents}.
\newblock In \emph{Proceedings of the 2018 Conference on Empirical Methods in Natural Language Processing}, pages 2775--2779, Brussels, Belgium. Association for Computational Linguistics.

\bibitem[{Mei et~al.(2024)Mei, Xie, Yuan, and Jackson}]{mei2024turing}
Qiaozhu Mei, Yutong Xie, Walter Yuan, and Matthew~O Jackson. 2024.
\newblock A turing test of whether ai chatbots are behaviorally similar to humans.
\newblock \emph{Proceedings of the National Academy of Sciences}, 121(9):e2313925121.

\bibitem[{Morin et~al.(2017)Morin, Parker, Stepler, and Mercer}]{morin2017police}
R~Morin, K~Parker, R~Stepler, and A~Mercer. 2017.
\newblock Police views, public views.

\bibitem[{Nasution and Onan(2024)}]{nasution2024chatgpt}
Arbi~Haza Nasution and Aytug Onan. 2024.
\newblock Chatgpt label: Comparing the quality of human-generated and llm-generated annotations in low-resource language nlp tasks.
\newblock \emph{IEEE Access}.

\bibitem[{Omiye et~al.(2023)Omiye, Lester, Spichak, Rotemberg, and Daneshjou}]{omiye2023large}
Jesutofunmi~A Omiye, Jenna~C Lester, Simon Spichak, Veronica Rotemberg, and Roxana Daneshjou. 2023.
\newblock Large language models propagate race-based medicine.
\newblock \emph{NPJ Digital Medicine}, 6(1):195.

\bibitem[{Park et~al.(2015)Park, Schwartz, Eichstaedt, Kern, Kosinski, Stillwell, Ungar, and Seligman}]{park2015automatic}
Gregory Park, H~Andrew Schwartz, Johannes~C Eichstaedt, Margaret~L Kern, Michal Kosinski, David~J Stillwell, Lyle~H Ungar, and Martin~EP Seligman. 2015.
\newblock Automatic personality assessment through social media language.
\newblock \emph{Journal of personality and social psychology}, 108(6):934.

\bibitem[{Park et~al.(2024)Park, Schoenegger, and Zhu}]{park2024diminished}
Peter~S Park, Philipp Schoenegger, and Chongyang Zhu. 2024.
\newblock Diminished diversity-of-thought in a standard large language model.
\newblock \emph{Behavior Research Methods}, pages 1--17.

\bibitem[{Pellert et~al.(2023)Pellert, Lechner, Wagner, Rammstedt, and Strohmaier}]{pellert2023ai}
Max Pellert, Clemens~M Lechner, Claudia Wagner, Beatrice Rammstedt, and Markus Strohmaier. 2023.
\newblock Ai psychometrics: Assessing the psychological profiles of large language models through psychometric inventories.
\newblock \emph{Perspectives on Psychological Science}, page 17456916231214460.

\bibitem[{{Pew Research Center}(2021)}]{pew2021demographics}
{Pew Research Center}. 2021.
\newblock Demographics and lifestyle differences among typology groups.

\bibitem[{{Pew Research Center}(2024)}]{pew2024immigration}
{Pew Research Center}. 2024.
\newblock How americans view the situation at the u.s.-mexico border, its causes and consequences.

\bibitem[{Roller et~al.(2021)Roller, Dinan, Goyal, Ju, Williamson, Liu, Xu, Ott, Smith, Boureau, and Weston}]{roller-etal-2021-recipes}
Stephen Roller, Emily Dinan, Naman Goyal, Da~Ju, Mary Williamson, Yinhan Liu, Jing Xu, Myle Ott, Eric~Michael Smith, Y-Lan Boureau, and Jason Weston. 2021.
\newblock \href {https://doi.org/10.18653/v1/2021.eacl-main.24} {Recipes for building an open-domain chatbot}.
\newblock In \emph{Proceedings of the 16th Conference of the European Chapter of the Association for Computational Linguistics: Main Volume}, pages 300--325, Online. Association for Computational Linguistics.

\bibitem[{Rottger et~al.(2022)Rottger, Vidgen, Hovy, and Pierrehumbert}]{rottger-etal-2022-two}
Paul Rottger, Bertie Vidgen, Dirk Hovy, and Janet Pierrehumbert. 2022.
\newblock \href {https://doi.org/10.18653/v1/2022.naacl-main.13} {Two contrasting data annotation paradigms for subjective {NLP} tasks}.
\newblock In \emph{Proceedings of the 2022 Conference of the North American Chapter of the Association for Computational Linguistics: Human Language Technologies}, pages 175--190, Seattle, United States. Association for Computational Linguistics.

\bibitem[{Sanderson et~al.(2021)Sanderson, Semyonov, and Gorodzeisky}]{sanderson2021declining}
Matthew~R Sanderson, Moshe Semyonov, and Anastasia Gorodzeisky. 2021.
\newblock Declining and splitting: Opposition to immigration in the united states, 1996--2018.
\newblock \emph{International Journal of Intercultural Relations}, 80:27--39.

\bibitem[{Santurkar et~al.(2023)Santurkar, Durmus, Ladhak, Lee, Liang, and Hashimoto}]{santurkar2023whose}
Shibani Santurkar, Esin Durmus, Faisal Ladhak, Cinoo Lee, Percy Liang, and Tatsunori Hashimoto. 2023.
\newblock Whose opinions do language models reflect?
\newblock In \emph{International Conference on Machine Learning}, pages 29971--30004. PMLR.

\bibitem[{Sap et~al.(2019)Sap, Card, Gabriel, Choi, and Smith}]{sap2019risk}
Maarten Sap, Dallas Card, Saadia Gabriel, Yejin Choi, and Noah~A Smith. 2019.
\newblock The risk of racial bias in hate speech detection.
\newblock In \emph{Proceedings of the 57th annual meeting of the association for computational linguistics}, pages 1668--1678.

\bibitem[{Sap et~al.(2022)Sap, Swayamdipta, Vianna, Zhou, Choi, and Smith}]{sap2022annotators}
Maarten Sap, Swabha Swayamdipta, Laura Vianna, Xuhui Zhou, Yejin Choi, and Noah~A Smith. 2022.
\newblock Annotators with attitudes: How annotator beliefs and identities bias toxic language detection.
\newblock In \emph{Proceedings of the 2022 Conference of the North American Chapter of the Association for Computational Linguistics: Human Language Technologies}, pages 5884--5906.

\bibitem[{Schwartz et~al.(2017)Schwartz, Giorgi, Sap, Crutchley, Ungar, and Eichstaedt}]{schwartz-etal-2017-dlatk}
H.~Andrew Schwartz, Salvatore Giorgi, Maarten Sap, Patrick Crutchley, Lyle Ungar, and Johannes Eichstaedt. 2017.
\newblock \href {https://doi.org/10.18653/v1/D17-2010} {{DLATK}: Differential language analysis {T}ool{K}it}.
\newblock In \emph{Proceedings of the 2017 Conference on Empirical Methods in Natural Language Processing: System Demonstrations}, pages 55--60, Copenhagen, Denmark. Association for Computational Linguistics.

\bibitem[{Serapio-García et~al.(2023)Serapio-García, Safdari, Crepy, Sun, Fitz, Romero, Abdulhai, Faust, and Matarić}]{serapiogarcia2023personality}
Greg Serapio-García, Mustafa Safdari, Clément Crepy, Luning Sun, Stephen Fitz, Peter Romero, Marwa Abdulhai, Aleksandra Faust, and Maja Matarić. 2023.
\newblock \href {https://arxiv.org/abs/2307.00184} {Personality traits in large language models}.
\newblock \emph{Preprint}, arXiv:2307.00184.

\bibitem[{Silver(2024)}]{pew202israel}
Laura Silver. 2024.
\newblock Younger americans stand out in their views of the israel-hamas war.

\bibitem[{Simmons(2023)}]{simmons2023moral}
Gabriel Simmons. 2023.
\newblock Moral mimicry: Large language models produce moral rationalizations tailored to political identity.
\newblock In \emph{Proceedings of the 61st Annual Meeting of the Association for Computational Linguistics (Volume 4: Student Research Workshop)}, pages 282--297.

\bibitem[{Tan et~al.(2024)Tan, Beigi, Wang, Guo, Bhattacharjee, Jiang, Karami, Li, Cheng, and Liu}]{tan2024large}
Zhen Tan, Alimohammad Beigi, Song Wang, Ruocheng Guo, Amrita Bhattacharjee, Bohan Jiang, Mansooreh Karami, Jundong Li, Lu~Cheng, and Huan Liu. 2024.
\newblock Large language models for data annotation: A survey.
\newblock \emph{arXiv preprint arXiv:2402.13446}.

\bibitem[{Thapa et~al.(2023)Thapa, Naseem, and Nasim}]{thapa2023humans}
Surendrabikram Thapa, Usman Naseem, and Mehwish Nasim. 2023.
\newblock From humans to machines: can chatgpt-like llms effectively replace human annotators in nlp tasks.
\newblock In \emph{Workshop Proceedings of the 17th International AAAI Conference on Web and Social Media}.

\bibitem[{T{\"o}rnberg(2023)}]{tornberg2023chatgpt}
Petter T{\"o}rnberg. 2023.
\newblock Chatgpt-4 outperforms experts and crowd workers in annotating political twitter messages with zero-shot learning.
\newblock \emph{arXiv preprint arXiv:2304.06588}.

\bibitem[{Uma et~al.(2021)Uma, Fornaciari, Hovy, Paun, Plank, and Poesio}]{uma2021learning}
Alexandra~N Uma, Tommaso Fornaciari, Dirk Hovy, Silviu Paun, Barbara Plank, and Massimo Poesio. 2021.
\newblock Learning from disagreement: A survey.
\newblock \emph{Journal of Artificial Intelligence Research}, 72:1385--1470.

\bibitem[{Veselovsky et~al.(2023)Veselovsky, Ribeiro, Cozzolino, Gordon, Rothschild, and West}]{veselovsky2023prevalence}
Veniamin Veselovsky, Manoel~Horta Ribeiro, Philip Cozzolino, Andrew Gordon, David Rothschild, and Robert West. 2023.
\newblock \href {https://arxiv.org/abs/2310.15683} {Prevalence and prevention of large language model use in crowd work}.
\newblock \emph{Preprint}, arXiv:2310.15683.

\bibitem[{Wake et~al.(2023)Wake, Kanehira, Sasabuchi, Takamatsu, and Ikeuchi}]{wake2023bias}
Naoki Wake, Atsushi Kanehira, Kazuhiro Sasabuchi, Jun Takamatsu, and Katsushi Ikeuchi. 2023.
\newblock Bias in emotion recognition with chatgpt.
\newblock \emph{arXiv preprint arXiv:2310.11753}.

\bibitem[{Wan et~al.(2023)Wan, Pu, Sun, Garimella, Chang, and Peng}]{wan2023kelly}
Yixin Wan, George Pu, Jiao Sun, Aparna Garimella, Kai-Wei Chang, and Nanyun Peng. 2023.
\newblock " kelly is a warm person, joseph is a role model": Gender biases in llm-generated reference letters.
\newblock \emph{arXiv preprint arXiv:2310.09219}.

\bibitem[{Zamarro et~al.(2020)Zamarro, Perez-Arce, and Prados}]{zamarro2020gender}
Gema Zamarro, Francisco Perez-Arce, and Maria~Jose Prados. 2020.
\newblock Gender differences in the impact of covid-19.
\newblock \emph{KTLA. Accessed on July}, 16:2021.

\bibitem[{Zhang et~al.(2018)Zhang, Dinan, Urbanek, Szlam, Kiela, and Weston}]{zhang-etal-2018-personalizing}
Saizheng Zhang, Emily Dinan, Jack Urbanek, Arthur Szlam, Douwe Kiela, and Jason Weston. 2018.
\newblock \href {https://doi.org/10.18653/v1/P18-1205} {Personalizing dialogue agents: {I} have a dog, do you have pets too?}
\newblock In \emph{Proceedings of the 56th Annual Meeting of the Association for Computational Linguistics (Volume 1: Long Papers)}, pages 2204--2213, Melbourne, Australia. Association for Computational Linguistics.

\bibitem[{Ziems et~al.(2024)Ziems, Held, Shaikh, Chen, Zhang, and Yang}]{ziems2024can}
Caleb Ziems, William Held, Omar Shaikh, Jiaao Chen, Zhehao Zhang, and Diyi Yang. 2024.
\newblock Can large language models transform computational social science?
\newblock \emph{Computational Linguistics}, 50(1):237--291.

\end{thebibliography}

\appendix



\section{Additional Language Results}
\label{sec:app lang features}

Table \ref{tab:ngram effect sizes} shows the effect sizes of the top ngrams shown in Figure \ref{fig:wordclouds}. Correlations with LIWC and the Moral Foundations dictionary are shown in Tables \ref{tab:liwc} and \ref{tab:moral foundations}, respectively. 

\begin{table*}[]
\centering
\resizebox{\textwidth}{!}{
\begin{tabular}{cccccccccccccc}
\cline{1-2} \cline{4-5}  \cline{7-8} \cline{10-11}  \cline{13-14}
Term & Effect Size &  & Term & Effect Size &  & Term & Effect Size &  & Term & Effect Size &  & Term & Effect Size \\ \cline{1-2} \cline{4-5}  \cline{7-8} \cline{10-11}  \cline{13-14}
i've & .368 &  & experiences & .996 &  & immigration & 2.58 &  & brutality & 2.22 &  & could & 2.81 \\
, i've & .366 &  & feelings & .921 &  & laws & 2.16 &  & police brutality & 1.94 &  & people & 2.39 \\
decades & .366 &  & have personal & .864 &  & security & 2.10 &  & systemic & 1.93 &  & if & 2.00 \\
seen & .326 &  & ai & .825 &  & illegal & 1.85 &  & and & 1.67 &  & like & 2.00 \\
witnessed & .303 &  & don't & .822 &  & immigration laws & 1.76 &  & racial profiling & 1.64 &  & someone & 1.99 \\
really & -.296 &  & modern & -.843 &  & rights & 1.88 &  & or & -1.38 &  & legalization can & -2.06 \\
think it's & -.299 & \textit{} & providers & -.852 & \textit{} & humane & -2.01 & \textit{} & can & -1.54 & \textit{} & public & -2.25 \\
think & -.300 &  & my & -.894 &  & climate & -2.29 &  & emergencies & -1.60 &  & this can & -2.32 \\
i think & -.321 &  & as a & -.916 &  & change & -2.39 &  & might & -1.95 &  & reduction in & -2.34 \\
its & -.334 & \textit{} & partner & -1.03 & \textit{} & climate change & -2.43 & \textit{} & caucasian & -2.10 & \textit{} & can & -3.59 \\ \cline{1-2} \cline{4-5}  \cline{7-8} \cline{10-11}  \cline{13-14}
\multicolumn{2}{c}{(a) Age} &  & \multicolumn{2}{c}{(b) Gender} &  & \multicolumn{2}{c}{(c) Political Idealology} &  & \multicolumn{2}{c}{(d) Race} &  & \multicolumn{2}{c}{(e) Substance Use} \\
\multicolumn{2}{c}{(Palestine)} &  & \multicolumn{2}{c}{(Parenting)} &  & \multicolumn{2}{c}{(Immigration)} &  & \multicolumn{2}{c}{(Policing)} &  & \multicolumn{2}{c}{(Legalization)} \\
\end{tabular}
}
\caption{\textbf{N-gram} associated with each human factor across their respective domains. We show the top five most positively (top five rows) and negatively (bottom five rows) associated with each dimension. Product moment correlations reported in (a), Cohen's d in all others. All association significant at a BH corrected significance level of $p<0.05$.}
\label{tab:ngram effect sizes}
\end{table*}

\begin{table*}[]
\centering
\resizebox{\textwidth}{!}{
\begin{tabular}{cccccccccccccc}
\cline{1-2} \cline{4-5}  \cline{7-8} \cline{10-11}  \cline{13-14}
Category & Effect Size &  & Category & Effect Size &  & Category & Effect Size &  & Category & Effect Size &  & Category & Effect Size \\ \cline{1-2} \cline{4-5}  \cline{7-8} \cline{10-11}  \cline{13-14} 
VISUAL & .324 &  & ADJ & .979 &  & POWER & 2.08 &  & TIME & 2.48 &  & LINGUISTIC & 3.95 \\
TIME & .324 &  & TONE NEG & .882 &  & RISK & 2.03 &  & TONE NEG & 2.18 &  & FUNCTION & 3.86 \\
FOCUSPAST & .199 &  & CULTURE & .841 &  & CULTURE & 1.32 &  & EMO NEG & 1.56 &  & VERB & 3.52 \\
REWARD & .179 &  & TECH & .840 &  & POLITIC & 1.23 &  & FOCUSPAST & 1.57 &  & PPRON & 3.32 \\
ARTICLE & .152 &  & EMO NEG & .818 &  & AUXVERB & 1.10 &  & ADJ & 1.46 &  & PRONOUN & 3.30 \\
CERTITUDE & -.239 &  & AFFILIATION & -.821 &  & SOCREFS & -1.33 &  & VERB & -1.59 &  & DRIVES & -1.36 \\
IPRON & -.276 & \textit{} & HOME & -.827 & \textit{} & SOCBEHAV & -1.75 & \textit{} & FOCUSFUTURE & -1.86 & \textit{} & CULTURE & -1.38 \\
COGPROC & -.278 &  & FAMILY & -.859 &  & MORAL & -1.80 &  & TENTAT & -2.00 &  & POWER & -1.42 \\
COGNITION & -.291 &  & MONEY & -.968 &  & SOCIAL & -1.81 &  & COGNITION & -2.25 &  & MONEY & -1.59 \\
INSIGHT & -.315 & \textit{} & ARTICLE & -1.04 & \textit{} & PROSOCIAL & -2.02 & \textit{} & COGPROC & -2.57 & \textit{} & LIFESTYLE & -1.76 \\

\cline{1-2} \cline{4-5}  \cline{7-8} \cline{10-11}  \cline{13-14}
\multicolumn{2}{c}{(a) Age} &  & \multicolumn{2}{c}{(b) Gender} &  & \multicolumn{2}{c}{(c) Political Idealology} &  & \multicolumn{2}{c}{(d) Race} &  & \multicolumn{2}{c}{(e) Substance Use} \\
\multicolumn{2}{c}{(Palestine)} &  & \multicolumn{2}{c}{(Parenting)} &  & \multicolumn{2}{c}{(Immigration)} &  & \multicolumn{2}{c}{(Policing)} &  & \multicolumn{2}{c}{(Legalization)} \\
\end{tabular}
}
\caption{\textbf{LIWC} categories associated with each human factor across their respective domains. We show the top five most positively (top five rows) and negatively (bottom five rows) associated with each dimension. Product moment correlations reported in (a), Cohen's d in all others. All association significant at a BH corrected significance level of $p<0.05$.}
\label{tab:liwc}
\end{table*}

\begin{table*}[]
\centering
\resizebox{\textwidth}{!}{
\begin{tabular}{cccccccccccccc}
\cline{1-2} \cline{4-5}  \cline{7-8} \cline{10-11}  \cline{13-14}
Category & Effect Size &  & Category & Effect Size &  & Category & Effect Size &  & Category & Effect Size &  & Category & Effect Size \\ \cline{1-2} \cline{4-5}  \cline{7-8} \cline{10-11}  \cline{13-14} 
 &  &  & FAIRNESSVICE & .398 &  & AUTHORITYVICE & 2.02 &  & HARMVICE & 2.14 &  & PURITYVIRTUE & .341 \\
 &  &  & INGROUPVICE & .279 &  & INGROUPVICE & 1.34 &  & FAIRNESSVICE & 1.58 &  & MORALITYGENERAL & .316 \\
 &  &  & PURITYVIRTUE & .064 &  & AUTHORITYVIRTUE & 1.23 &  & INGROUPVIRTUE & .738 &  & PURITYVICE & .251 \\
 &  &  & AUTHORITYVICE & -.006 &  & PURITYVICE & 1.08 &  & FAIRNESSVIRTUE & .610 &  & AUTHORITYVICE & 0.204 \\
 &  &  & AUTHORITYVIRTUE & -.084 &  & HARMVIRTUE & .881 &  & HARMVIRTUE & .544 &  & AUTHORITYVIRTUE & -.301 \\
 &  &  & FAIRNESSVIRTUE & -.137 &  & PURITYVIRTUE & .555 &  & PURITYVIRTUE & .399 &  & HARMVIRTUE & -.352 \\
 &  & \textit{} & HARMVIRTUE & -.347 & \textit{} & MORALITYGENERAL & .386 & \textit{} & MORALITYGENERAL & .174 & \textit{} & FAIRNESSVIRTUE & -.465 \\
 &  &  & INGROUPVIRTUE & -.367 &  & INGROUPVIRTUE & -.066 &  & AUTHORITYVICE & .154 &  & INGROUPVIRTUE & -1.05 \\
MORALITYGENERAL & -.185 &  & HARMVICE & -.494 &  & HARMVICE & -1.07 &  & INGROUPVICE & -.395 &  & INGROUPVICE & -1.06 \\
FAIRNESSVIRTUE & -.232 & \textit{} & MORALITYGENERAL & -.527 & \textit{} & FAIRNESSVIRTUE & -1.32 & \textit{} & AUTHORITYVIRTUE & -.735 & \textit{} & FAIRNESSVICE & -1.154 \\
\cline{1-2} \cline{4-5}  \cline{7-8} \cline{10-11}  \cline{13-14}
\multicolumn{2}{c}{(a) Age} &  & \multicolumn{2}{c}{(b) Gender} &  & \multicolumn{2}{c}{(c) Political Idealology} &  & \multicolumn{2}{c}{(d) Race} &  & \multicolumn{2}{c}{(e) Substance Use} \\
\multicolumn{2}{c}{(Palestine)} &  & \multicolumn{2}{c}{(Parenting)} &  & \multicolumn{2}{c}{(Immigration)} &  & \multicolumn{2}{c}{(Policing)} &  & \multicolumn{2}{c}{(Legalization)} \\
\end{tabular}
}
\caption{\textbf{Moral Foundations} categories associated with each human factor across their respective domains. We show the top five most positively (top five rows) and negatively (bottom five rows) associated with each dimension. Product moment correlations reported in (a), Cohen's d in all others. All association significant at a BH corrected significance level of $p<0.05$.}
\label{tab:moral foundations}
\end{table*}

\section{Algorithm}
\label{sec:app algorithm}

Algorithm \ref{alg: correlations} shows how the correlations in Tables \ref{tab:explicit cross correlations}, \ref{tab:explicit vs implicit}, and \ref{tab:importance} are calculated. The algorithm shows an example of comparing a full persona to a single dimension persona (gender) in the parenting domain, but, in general, this algorithm takes in two personas and a domain. For the convergent / divergent validity tests (Table \ref{tab:explicit cross correlations}) we consider all explicit, single factor persona pairs across all domains. For the implicit vs explicit analysis (Table \ref{tab:explicit vs implicit}), we consider one explicit and one implicit persona, across all pairs, and across all domains. Finally, for the persona importance task (Table \ref{tab:importance}) we consider a full persona and a single factor persona, for all human factors, and across all domains. 

\begin{algorithm}[t]
\caption{Extracting Word Frequencies and Calculating Correlations}
\label{alg:word_frequencies}
\begin{algorithmic}
\State \textbf{Function:} \textit{LLM(persona\_type, task)}: A function that prompts an LLM with a persona type (\textit{full} or \textit{individual}) for a generation task
\State

\State $gen\_f \gets [LLM(\text{full}, \text{parenting})\text{ for each }$
\State \hspace{25mm}$\text{ worker}]$
\State $\textit{all\_words} \gets \{ w_i : \text{frequency of } w_i \text{ in } gen\_f \}$
\State $corr_1 \gets [\quad]$
\State $gender \gets [\text{gender labels for each worker}]$
\State

\For{each $w_i$ in \textit{all\_words}}
    \State $r \gets \text{corr}(\Sigma(w_i), gender)$
    \State $\textit{corr\_1.append}(r)$
\EndFor
\State

\State $gen\_bi \gets LLM(\text{gender}, \text{parenting})$
\State $\textit{all\_words} \gets \{ w_i : \text{frequency of } w_i \text{ in } gen\_bi \}$
\State $corr_2 \gets [\quad]$
\State

\For{each $w_i$ in \textit{all\_words}}
    \State $r \gets \text{corr}(\Sigma(w_i), gender)$
    \State $\textit{corr\_2.append}(r)$
\EndFor
\State

\State $importance \gets \text{corr}(corr_1, corr_2)$

\end{algorithmic}
\label{alg: correlations}
\end{algorithm}

\section{Belief Generation Validation}
\label{sec:validation}

Table \ref{tab:validation questions} shows the questions given to the LLM in the validation analysis for \textbf{BGT1}. These questions were designed to be binary versions of the beliefs outlined in Table \ref{tab:beliefs}.

\begin{table*}[h]
\resizebox{.98\textwidth}{!}{%
\begin{tabular}{ll p{8cm}}
\toprule
Human Factor & Domain & Question \\ \midrule
Age & Palestine & Does the following text indicate that the writer’s sympathies lie more with the Israeli people (1) or more with the Palestinian people (-1) or both (0)? \\
Gender & Parenting & Does the following text indicate that the writer feels a great deal of pressure to focus on their responsibilities at home(1), work (-1), or both (0)? \\
Political Ideology & Immigration & Does the following text indicate that the writer feels that there should be stricter (1) or looser (0) restrictions on immigration at the U.S. border with Mexico? \\
Race & Policing & Does the following text indicate that the writer sees the police as protectors (0), enforces (-1), or both (0)?  \\
Substance Use & Legalization & Does the following text indicate that the writer supports legalization of drugs? Yes (1), No (-1), or neither (0) \\ \bottomrule
\end{tabular}
}
\caption{Questions used in the Belief Generation Validation task.}
\label{tab:validation questions}
\end{table*}

\section{Annotation Reliability}
\label{sec:reliability}

Here we calculate pairwise Fleiss kappa's for each combination of human, explicit Persona-LLMs, and implicit Persona-LLMs, matching the analysis in \citet{bavaresco2024llms}. Using Human vs. Explicit as an example, we calculate Fleiss kappa ($\kappa$) between the ratings (on the 5 posts) of each of the 641 humans and the ratings of each of the 641 explicit Persona-LLMs, for a given persona type (e.g., female). This results in $641^2$ kappas, which we then average and report in Table \ref{tab:reliability}. Here we see that explicit Persona-LLMs and implicit Persona-LLMs tend to agree more than humans and either type of Persona-LLM.

\begin{table}[]
\resizebox{\columnwidth}{!}{
\begin{tabular}{@{}lccc@{}}
\toprule
 & \begin{tabular}[c]{@{}c@{}}Human\\ vs.\\ Explicit\end{tabular} & \begin{tabular}[c]{@{}c@{}}Human\\ vs.\\ Implicit\end{tabular} & \begin{tabular}[c]{@{}c@{}}Explicit\\ vs.\\ Implicit\end{tabular} \\ \midrule
All & .42 & .43 & .76 \\
Black & .39 & .45 & .61 \\
White & .43 & .43 & .78 \\
Female & .43 & .43 & .76 \\
Male & .43 & .43 & .75 \\
Conservative & .40 & .40 & .77 \\
Liberal & .45 & .45 & .77 \\
Uses Substances & - & - & .78 \\
No Substances & - & - & .76 \\ \bottomrule
\end{tabular}
}
\caption{Average pairwise Fleiss kappa's for each combination of persona type across humans, explicit Persona-LLMs, and implicit Persona-LLMs.}
\label{tab:reliability}
\end{table}

\end{document}